\PassOptionsToPackage{table}{xcolor}
\documentclass[10pt,twocolumn,letterpaper]{article}
\usepackage[review]{cvpr}

\usepackage{cuted} 

\usepackage{currfile} 

\usepackage{caption} 

\definecolor{cvprblue}{rgb}{0.21,0.49,0.74}
\usepackage[pagebackref,breaklinks,colorlinks,allcolors=cvprblue]{hyperref}
\usepackage{graphicx}
\definecolor{cvprblue}{rgb}{0.21,0.49,0.74}
\usepackage[pagebackref,breaklinks,colorlinks,allcolors=cvprblue]{hyperref}
\definecolor{sh_gray}{rgb}{0.84,0.84,0.84}
\definecolor{sh_gray2}{rgb}{1,0.89,0.75}
\definecolor{color3}{rgb}{0.95,0.95,0.95}

\definecolor{color4}{rgb}{0.94,0.94,1}
\definecolor{color5}{rgb}{0.90,0.90,0.90}
\def\confName{CVPR}
\def\confYear{2027}

\title{DeepRFTv3: Deep Residual Fourier Convolution for Unified Image Deblurring}
\author{Xintian~Mao$^{1}$,
        Haofei Song$^{1}$, 
        Huaiyuan Wang$^{1}$, 
        Jiansheng Wang$^{1,2}$,
        Qingli~Li$^{1}$,
        Yan~Wang$^{1*}$  \\ 
        $^{1}$ Shanghai Key Laboratory of Multidimensional Information Processing, East China Normal University, China.\\ 
        $^{2}$ Shanghai Institute of Technical Physics, Chinese Academy of Sciences, China. \\
        $^{*}$ Corresponding Author \{ywang@cee.ecnu.edu.cn\ \\
        \small Code: \url{https://github.com/DeepMed-Lab-ECNU/Single-Image-Deblur}}
\begin{document}
\twocolumn[{%
\renewcommand\twocolumn[1][]{#1}%
\maketitle
\begin{center}
    \centering
    \captionsetup{type=figure}
    \vspace{-1.em}    \includegraphics[width=0.96\linewidth]{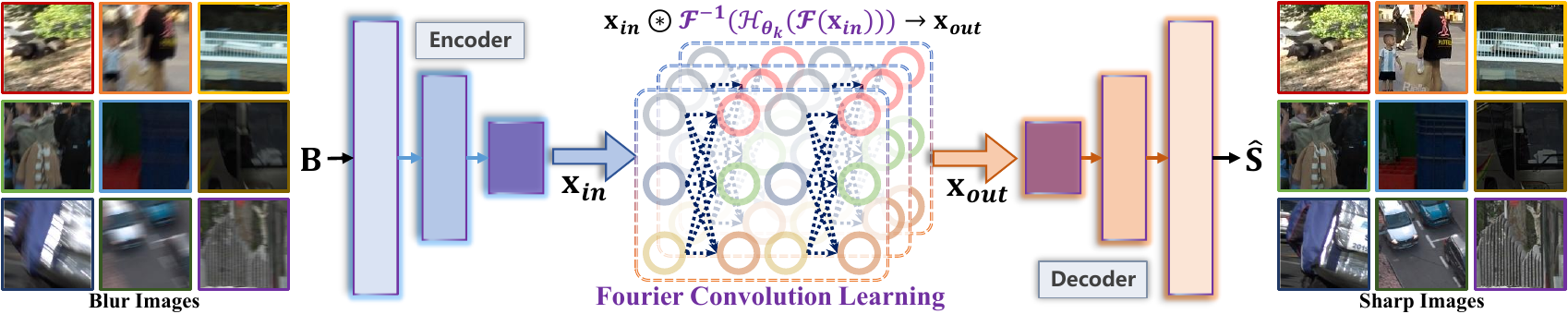}
    \vspace{-1.em}
    \caption{
    \label{fig:1}
    Deep Residual Fourier Convolution for Unified Image Deblurring.} 
\end{center}%
}]

\begin{abstract}

Image deblurring is a fundamental task in computer vision, yet existing methods mainly specialize in isolated scenarios. Consequently, the deblurring performance of these models drops significantly across scenarios with different cameras and image contents. This motivates the development of a unified motion deblurring framework that can handle diverse blur patterns. To this end, we construct a multi-source dataset and develop a data pruning and quality control pipeline. Then, to restore blur images with various motion blur kernels across different scenarios, we propose a unified motion deblurring framework via Fourier Convolution Learning. Existing deblurring networks usually rely on pixel-level mappings and overlook the convolutional nature of blur. Thus, the network mostly fits the image content itself rather than learning the convolution process, making it difficult to handle multi-source blur images. Instead of relying on spatial mapping, we perform blur-adaptive global filter learning in Fourier space, which avoids the limited spatial support of kernel estimation. Extensive experiments demonstrate that our unified model effectively handles motion blurs across diverse sources and scenes, achieving competitive performance across both in-domain and OOD benchmarks.
\end{abstract}
\section{Introduction}
\label{sec:intro}
Image deblurring, as a fundamental subtask of image restoration, aims to recover latent sharp images from blur inputs~\cite{nah2017deep, kupyn2019deblurgan}. In real-world scenarios, the formation of a blurry image $\mathrm{B}$ is formulated as the spatial convolution of a sharp latent image $\mathrm{S}$ with a motion blur kernel $\mathrm{k}_{blur}$, along with additive noise $\mathrm{n}$~\cite{fergus2006removing, pan2016blind}:
\begin{equation}
\mathrm{B} = \mathrm{S} \circledast \mathrm{k}_{blur} + \mathrm{n},
\label{eq:blur}
\end{equation}
where $\circledast$ denotes the 2D convolution operator. This degradation process destroys high-frequency edge details and delicate textures, leading to visually unpleasant artifacts and severely impairing downstream vision tasks.

Recently, deep learning-based approaches have achieved remarkable improvements in image deblurring~\cite{zamir2021multi, chen2022simple, wang2022uformer}. Existing methods predominantly develop specialized frameworks on isolated benchmark datasets. However, as shown in Table~\ref{tab:summary_datasets}, real-world motion blurs exhibit substantial divergence across datasets due to disparate cameras (GoPro, Sony, CHRONOS), distinct resolutions, and diverse scenes (outdoor streetscapes, pedestrians, night-time, low-light). Consequently, models trained on a single isolated scenario suffer from severe performance drops when deployed across unseen environments. To alleviate cross-domain degradation, recent All-in-One restoration models attempt to handle diverse corruptions within a unified framework~\cite{li2022AirNet, potlapalli2023promptir}. Nonetheless, their performance on motion deblurring remains constrained. Denoising and deraining are predominantly additive corruptions, whereas motion blur is a convolutional degradation involving structural distortion and phase shifts; forcibly optimizing these tasks within a single model inevitably causes optimization conflicts and feature interference. Moreover, these All-in-One approaches~\cite{potlapalli2023promptir, tang2026BaryIR} heavily over-rely on the synthetic GoPro dataset~\cite{nah2017deep}, which comprises only 3.2K blur-sharp pairs across 33 scenes (with merely 2.1K training pairs). Models trained on such limited distributions lack the rich semantic and kernel variations of the real world, leading to severe overfitting. To this end, developing a unified motion deblurring framework becomes highly imperative.

\begin{figure*}[t]
\begin{center}
    \includegraphics[width=1.\linewidth]{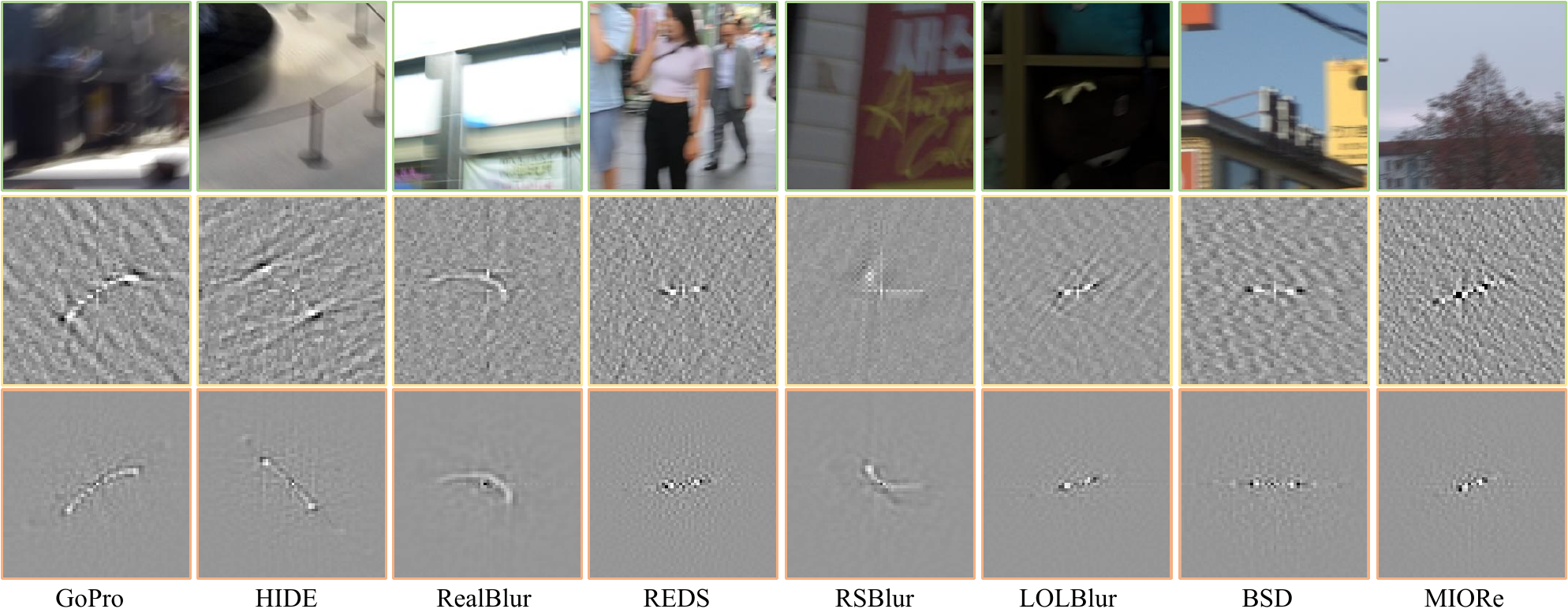}
\end{center}
\vspace{-1.2em}
\caption{Illustration of DeepRFTv3 and visualization of blur images, deconvolution kernels and kernels by FCL on various datasets.}
\label{fig:kernels}
\vspace{-1.em}
\end{figure*}

\begin{table*}[t]
\centering

\caption{Summary of various motion deblurring datasets.}
\label{tab:summary_datasets}
\vspace{-0.5em}
\resizebox{\textwidth}{!}{
\begin{tabular}{l|cccccc}
\toprule
Dataset & Acquisition & Images & Videos & Camera &Resolution & Content \\
\midrule
GoPro~\cite{nah2017deep} & Averaged & 3.2K & 33  & GoPro HERO4 & 1280 $\times$ 720 & Outdoor Scenes\\
HIDE~\cite{Shen2019human} & Averaged & 8.4K & 31  & GoPro HERO & 1280 $\times$ 720 & Pedestrian in Outdoor  \\
REDS~\cite{Nah2021ntire} & Averaged & 30K & 300  & GoPro HERO6 & 1280 $\times$ 720 & Outdoor Scenes\\
RSBlur~\cite{rim2022RSBlur} & Beam Splitter & 13.3K & 697 & - & $\approx$1915 $\times$ 1200 & Outdoor Scenes\\
RealBlur~\cite{Rim2020real} & Beam Splitter & 9.1K & 232  & Sony A7RM3 & $\approx$660 $\times$ 750 & Static Object at Night \\
LOLBlur~\cite{zhou2022lednet} & Simulated & 12K & 200  & Sony RX10 IV & 1120 $\times$ 640 & Low-light Scenes \\
MIORe~\cite{ciubotariu2025MIORe} & Averaged & 135K & 333  & CHRONOS 2.1-HD & 1920 $\times$ 1080 & Outdoor Scenes \\
FoundIR~\cite{li2025foundir} & Slider & - & - & Different Cameras  & 1920 $\times$ 1080 & Static Object \\
BSD~\cite{zhong2023BSD} & Beam Splitter & 33K & 100  & - & 640 $\times$ 480 & Outdoor Scenes\\
\arrayrulecolor{black}\bottomrule
\end{tabular}
}
\vspace{-1.3em}
\end{table*}

As shown in Fig.~\ref{fig:kernels}, we present a unified motion deblurring framework, termed \textbf{DeepRFTv3}, driven by \textbf{Fourier Convolution Learning} (FCL). On the data side, recognizing that high-quality datasets are essential for reliable training and evaluation, we design a multi-source training/evaluation protocol equipped with a dedicated data pruning and quality control pipeline. This pipeline removes samples with obvious quality issues, such as spatial misalignment and residual blur in the ground truth, to prevent the network from fitting erroneous degradation patterns. Consequently, it ensures clean supervision for model optimization and reduces the influence of obvious acquisition and annotation artifacts across both in-domain and out-of-domain (OOD) benchmarks. 

On the framework side, mainstream networks are typically confined to spatial pixel-to-pixel black-box mapping, often overlooking the convolutional nature of motion blur. To overcome this limitation, we elevate the restoration paradigm to blur-adaptive convolution learning. By exploiting the Convolution Theorem, FCL reformulates spatial kernel estimation as element-wise multiplication in the Fourier space. As shown in Fig.~\ref{fig:kernels}, this formulation enables the network to predict blur-adaptive frequency filters and dynamically modulate the deep features. Thus, it captures degradation-aware kernels in diverse motion blur scenes with minimal computational overhead.

The main contributions are summarized as follows:
\begin{itemize}
    \item We propose DeepRFTv3, a unified motion deblurring framework with multi-source training and evaluation, ensuring effective training supervision and evaluation.
    \item We design the Fourier Convolution Learning module, which reformulates complex spatial kernel prediction into efficient frequency-domain multiplication, enabling blur-adaptive filter learning to effectively capture diverse input-specific blurs.
    \item Extensive experiments on both in-domain and OOD benchmarks demonstrate that our method achieves competitive restoration performance and superior cross-domain generalization ability.
\end{itemize}
\section{Related Work}
\label{sec:related}

\noindent\textbf{Deep Image Deblurring}
In recent years, deep learning-based methods have achieved significant success in image deblurring~\cite{nah2017deep, tao2018scale, zamir2021multi}. Pioneering approaches attempted to estimate the blur kernel using Convolutional Neural Networks (CNNs) and then restore the latent image via non-blind deconvolution~\cite{sun2015learning}. However, these methods struggle with spatially-variant blurs in dynamic scenes and are sensitive to noise. Consequently, almost all recent methods have shifted towards end-to-end training paradigms, mapping blurry images directly to their sharp counterparts. To achieve better performance, various architectural improvements have been proposed, ranging from multi-stage CNNs (MPRNet~\cite{zamir2021multi}, DMPHN~\cite{Zhang2019DMPHN}) to advanced Vision Transformers (Uformer~\cite{wang2022uformer}, Restormer~\cite{Zamir2021restormer}). Despite their success, these methods primarily focus on \textit{pixel-level} restoration, merely learning the residual between blur and sharp image pairs while overlooking the convolutional nature of the blur process. To revisit kernel learning, some recent works introduce kernel prior-assisted models: UFPNet~\cite{fang2023UFPNet} pre-trains a flow-based model to estimate spatially variant kernel maps, and PGDN~\cite{Fang_PGDN} introduces a parameterized motion kernel estimator. However, kernels learned in the spatial domain are often limited in size and capacity.


\noindent\textbf{Frequency Domain Methods for Vision Tasks}
A growing literature corpus has demonstrated the effectiveness of extracting information from the frequency domain for various vision tasks~\cite{rippel2015spectral, ZequnQin2020FcaNet, rao2021global}. Since the Convolution Theorem states that convolution in the spatial domain is equivalent to element-wise multiplication in the frequency domain, Fourier space presents a tractable framework for handling large-scale spatial transformations. For example, LaMa~\cite{Suvorov2022resolution} applies Fast Fourier Convolution (FFC) to capture a wider receptive field for image inpainting. GFNet~\cite{rao2021global} learns long-term spatial dependencies in the frequency domain for image classification. For image deblurring, DeepRFT~\cite{XintianMao2023DeepRFT} introduces an additive paradigm of Res-FFT-ReLU to recognize blur patterns. LoFormer~\cite{mao2024Loformer} and FFTformer~\cite{kong2023fftformer} further explore frequency information for self-attention. SegDeblur~\cite{kim2024SegDeblur} and Self-Deblur~\cite{kim2025refdeblur} try to predict the image-level deconvolution kernel in logarithmic Fourier space for reliable prior information.
However, these approaches merely utilize frequency representations for additive feature aggregation or auxiliary prior, leaving the potential of convolution learning in the feature space largely underexplored for the deblurring task.

\noindent\textbf{Unified Image Restoration}
Nowadays, almost all existing deblurring models are trained and evaluated on specific datasets (GoPro~\cite{nah2017deep} for global camera shake, LOLBlur~\cite{zhou2022lednet} for night-time blur). This reliance on single-dataset training severely limits their generalization capability when deployed in complex, real-world open scenarios. Recently, All-in-One image restoration models~\cite{li2022AirNet, potlapalli2023promptir, hu2025dcpt, tang2026BaryIR} attempted to solve multiple degradations (such as noise, rain, and blur) using a single unified network. However, denoising and deraining are additive degradations requiring local spatial aggregation, whereas motion blur is a convolutional degradation that demands large-scale deformation and phase shift correction, so that their performance on deblurring remains constrained.

DeMoE~\cite{feijoo2026DeMoE} proposed a Mixture-of-Experts decoder for unified image deblurring, aiming to address both motion and defocus blur within a single model. However, compared to the diverse range of available motion blur benchmarks, existing defocus datasets~\cite{Abuolaim2020DPDD,Lee2021IFAN_RealDOF,ruan2021LFDOF,li2023SDD} suffer from several inherent issues, such as color shifts, pixel misalignment, and limited scene diversity. Therefore, instead of mixing multiple degradations, we specifically target a {Unified Motion Deblurring} framework which learns the essence of blur degradation via {Fourier Convolution Learning}.


\begin{figure*}[t]
\begin{center}
    \includegraphics[width=1.\linewidth]{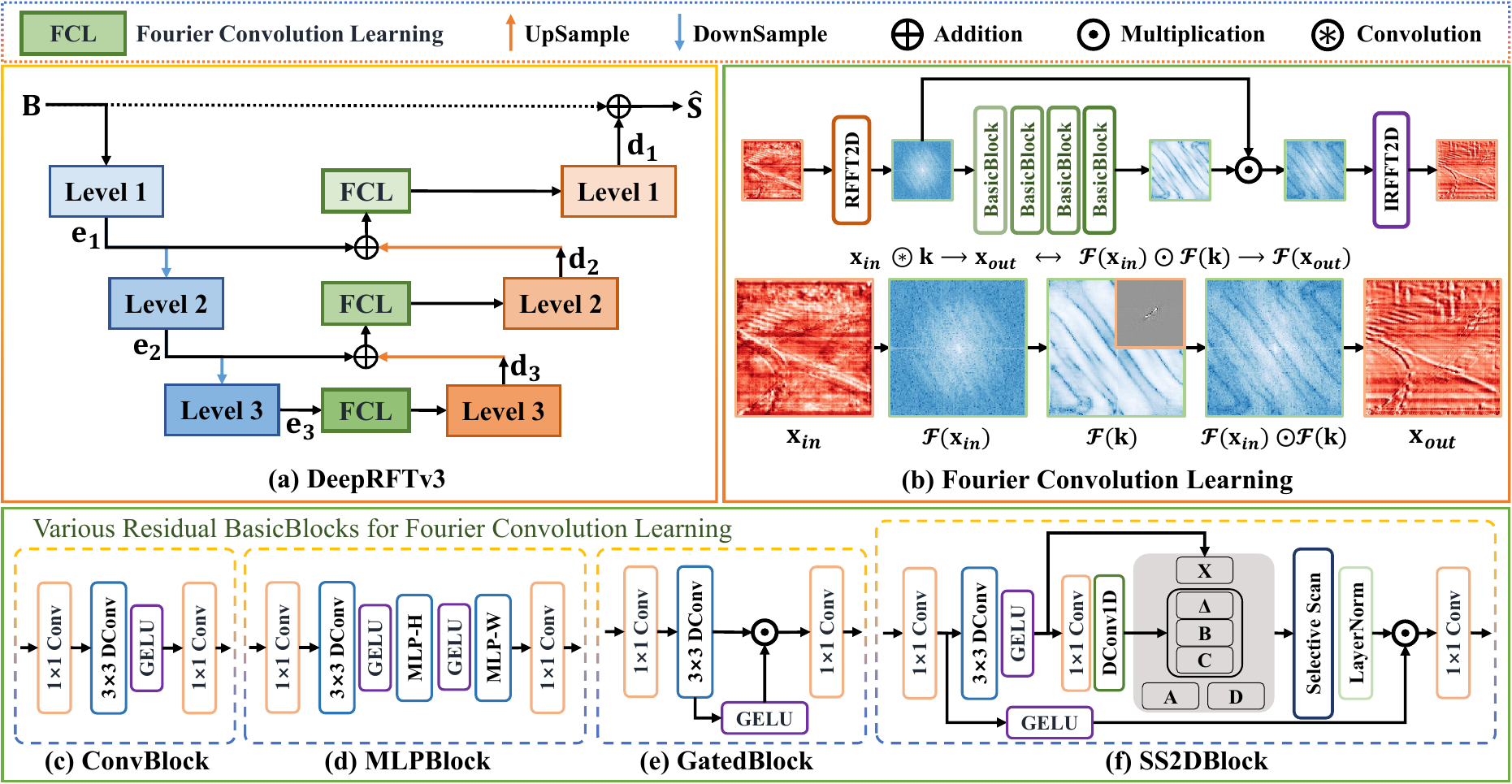}
\end{center}
\vspace{-1.em}
\caption{Architecture of DeepRFTv3, Fourier Convolution Learning and Basicblocks.}
\label{fig:arch}
\end{figure*}

\section{Unified Motion Deblurring Framework}
\noindent\textbf{Overall Architecture.} As illustrated in Fig.~\ref{fig:arch}(a), the overall architecture of our DeepRFTv3Net adopts a 3-level U-Net structure, inspired by the recent success of state-of-the-art architectures such as EVSSM~\cite{kong2025EVSSM} and BioIR~\cite{cui2026BioIR}. The restoration process can be formally described as follows:
\begin{align}
\mathrm {B} \xrightarrow{\mathcal{E}_{\theta_{e}}}\left \{  \mathrm{e}_{1}, \mathrm{e}_{2}, \mathrm{e}_{3} \right \} \xrightarrow{\mathcal{D}_{\theta_{d}} }
\left \{ \mathrm{d}_{1}, \mathrm{d}_{2}, \mathrm{d}_{3} \right \} \xrightarrow{\mathcal{T}_{\theta_{t}}} \mathrm {\hat{P}}\xrightarrow{+\mathrm {B}} \mathrm{\hat{S}},
\end{align}
where $\mathcal{E}_{\theta_{e}}$, $\mathcal{D}_{\theta_{d}}$ and $\mathcal{T}_{\theta_{t}}$ denote the encoder, the decoder and the convolution tail (parameterized by ${\theta_{e}}$, $\theta_{d}$ and $\theta_{t}$), $\left \{  \mathrm{e}_{1}, \mathrm{e}_{2}, \mathrm{e}_{3} \right \}$, $\left \{  \mathrm{d}_{1}, \mathrm{d}_{2}, \mathrm{d}_{3} \right \}$ and $\mathrm {\hat{P}}$ are the outputs of the encoder, the decoder and the convolution tail, $\mathrm{\hat{S}}$ is the restored image of the model.

\subsection{Fourier Convolution Learning}

\noindent\textbf{From Spatial Convolution to Frequency Filtering.}
As formulated in Eq.~\ref{eq:blur}, motion blur is fundamentally a convolution process. However, most existing deep networks rely on a direct pixel-to-pixel mapping strategy $\mathrm{B} \rightarrow \hat{\mathrm{S}}$ in the spatial domain. This black-box mapping merely attempts to fit spatial differences while overlooking the convolutional nature of the blur degradation. Forcing standard CNNs to implicitly learn complex, large-scale inverse convolutions inevitably leads to severe receptive field bottlenecks and optimization difficulties. To this end, we shift the convolution learning process to the Fourier space. According to the Convolution Theorem, spatial convolution is mathematically equivalent to element-wise multiplication in the frequency domain:
\begin{align}
    \mathrm{S} &= \mathcal{F}^{-1}\left( \mathcal{F}(\mathrm{B}) \odot \mathcal{F}(\mathrm{k}_{deblur}) \right),
    \label{eq:fourier_blur}
\end{align}
where $\mathrm{k}_{deblur}$ is the inverse kernel of $\mathrm{k}_{blur}$, $\mathcal{F}$ denotes the 2D Real Fast Fourier Transform (RFFT2D), $\mathcal{F}^{-1}$ is the Inverse RFFT2D, and $\odot$ is the Hadamard product. This conversion simply reformulates the intractable spatial kernel estimation into a simpler multiplication.

\noindent\textbf{Fourier Convolution Learning (FCL).}
Existing frequency-based architectures fail to fully exploit this property. For instance, DeepRFT~\cite{XintianMao2023DeepRFT} adopts addition:
\begin{equation}
    \mathrm{x}_{out} = \mathcal{F}^{-1}\left( \mathcal{F}(\mathrm{x}_{in}) + \mathcal{H}_{{\theta}}(\mathcal{F}(\mathrm{x}_{in})) \right),
\end{equation}
which treats frequency representations merely as additive residual features rather than convolution kernels. Conversely, while GFNet~\cite{rao2021global} utilizes frequency multiplication:
\begin{equation}
    \mathrm{x}_{out} = \mathcal{F}^{-1}\left( \mathcal{F}(\mathrm{x}_{in}) \odot \mathrm{W} \right),
\end{equation}
it learns a static global filter $\mathrm{W}$ that acts merely as a degradation-agnostic prompt, lacking the adaptability to handle varying, image-specific motion blurs.

Inspired by Eq.~\ref{eq:fourier_blur}, we propose the FCL module (Fig.~\ref{fig:arch}(b)) to instantiate a blur-adaptive convolution:
\begin{equation}
    \mathrm{x}_{out} = \mathcal{F}^{-1}\left( \mathcal{F}(\mathrm{x}_{in}) \odot \mathcal{H}_{{\theta}_k}(\mathcal{F}(\mathrm{x}_{in})) \right).
    \label{eq:fcl}
\end{equation}
Specifically, FCL transforms $\mathrm{x}_{in}$ from spatial domain to frequency domain, where a lightweight non-linear network $\mathcal{H}_{{\theta}_k}$ dynamically predicts blur-adaptive global filters $\mathcal{F}(\mathrm{k})$. These blur-adaptive filters are then multiplied with $\mathcal{F}(\mathrm{x}_{in})$ before applying the IRFFT2D. 

\noindent\textbf{Multi-Scale FCL.}
In standard U-Net architectures, encoder-decoder transitions are typically achieved via additive mechanisms (skip-addition): $\mathrm{\hat{x}}_i = \mathrm{e}_{i} + \mathrm{d}_{i-1}^\uparrow$. 
While effectively preserving high-frequency structures, this straightforward spatial combination simultaneously propagates redundant, degradation-corrupted features from the encoder. These degraded features can interfere with the decoder when reconstructing sharp details.

To address this issue, we introduce Multi-Scale FCL at all encoder-decoder transitions:
\begin{align}
    \mathrm{x}_i = \mathcal{F}^{-1}\left( \mathcal{F}(\mathrm{\hat{x}}_i) \odot \mathcal{H}_{{\theta}_k}(\mathrm{\mathcal{F}(\mathrm{\hat{x}}}_i)) \right),
    \label{eq:msfcl}
\end{align}
where the feature $\mathrm{\hat{x}}_i$ is refined into $\mathrm{x}_i$ before entering the decoder. By executing blur-adaptive filtering, the network effectively modulates the encoded representations and filters out degradation-irrelevant information. Consequently, Multi-Scale FCL transforms conventional skip connections into blur-adaptive information transmission, ensuring the decoder receives purified representations tailored to the specific blur pattern of the current input.

\begin{table*}[t!]
\footnotesize
\begin{center}
\vspace{-1.em}
\caption{Comparisons on in-domain datasets. }
\label{tab:test1}
\vspace{-0.7em}
\setlength{\tabcolsep}{1pt}
\resizebox{1\linewidth}{!}{
\begin{tabular}{l| ccc | ccc | ccc | ccc | cc | cc  }
\toprule[0.15em]
\multicolumn{1}{c|}{} & \multicolumn{3}{c|}{\textbf{GoPro}} & \multicolumn{3}{c|}{\textbf{HIDE}}& \multicolumn{3}{c|}{\textbf{REDS}} & \multicolumn{3}{c|}{\textbf{RSBlur}}& \multicolumn{2}{c|}{\textbf{RealBlur}}& \multicolumn{2}{c}{\textbf{LOLBlur}} \\
 \textbf{Method} & PSNR&{SSIM}&MUSIQ & PSNR&{SSIM}&MUSIQ& PSNR&{SSIM}&MUSIQ& PSNR&{SSIM}&MUSIQ& PSNR&{SSIM}& PSNR &{SSIM}\\
 \midrule[0.15em]
 GT & - & 1.000 &47.97& -& 1.000 &60.43& -&1.000 &66.37& -& 1.000 &37.31 & -& 1.000 & -& 1.000 \\
Blur &25.64 &0.793 &24.87 &23.94 &0.759 &26.28 &26.34 &0.772 &39.01 &28.74 &0.850 &21.90 &30.08 &0.874 &33.82 &0.904\\
\arrayrulecolor{black!30}\midrule
MIMOUNet &30.47  &0.906  &39.91  &28.93  &0.888  &47.22  &32.14  &0.908  &62.08 &32.30  &0.917  &30.57 &33.01  &0.925  &41.70  &0.977 \\
DeepRFT &31.92 &0.932 &41.11 &30.84 &0.921 &51.70 &32.36 &0.912 &61.21 &33.45  &0.937  &32.40&33.79 &0.938 &43.38  &0.982  \\
NAFNet &32.81 &0.940 &43.61 &31.11 &0.924 &54.99 &33.66 &0.931 &64.50 &33.89 &0.941 &34.62&33.93 &0.939 &44.22 &0.985  \\
LoFormer &32.33 &0.936 &42.28 &30.98 &0.922 &53.09 &33.62 &0.934 &64.34 &33.46 &0.937 &32.95&33.86 &0.939 &44.12 &0.985 \\
PromptIR &31.94 &0.931 &42.86 &30.95 &0.922 &54.85 &32.98 &0.922 &64.70 &33.65 &0.939&33.89&33.92 &0.939&43.84 &0.984\\
MoCEIR &32.36 &0.936 &43.54&31.19&0.925&55.53 &33.24 &0.926 &65.14 &33.48 &0.937&34.06&33.93&0.939&44.23 &0.985\\
BioIR &33.12 &0.943 &\textcolor{cyan}{44.77}&{31.77}&{0.933}&\textcolor{cyan}{57.34}&34.48 &0.942 &\textcolor{cyan}{66.00} &33.87&0.941&\textcolor{cyan}{35.22}&{34.05}&0.940&\textcolor{cyan}{45.11} &\textcolor{cyan}{0.987}\\
EVSSM&{33.26}&{0.945}&44.45&31.70&0.932&56.66&\textcolor{cyan}{34.68}&\textcolor{cyan}{0.944}&65.81 &{33.91} &{0.942} &34.95&34.04 &{0.940}&44.94 &{0.987}\\
\arrayrulecolor{black!30}\midrule
DeepRFTv2 &33.67  &0.949  &45.09  &32.08  &0.935  &57.54  &34.72  &0.944  &66.02 &34.28  &0.945  &35.97 &34.14  &0.942  &45.42  &0.988 \\
DeepRFTv3-B &\textcolor{cyan}{33.48}&\textcolor{cyan}{0.945}&44.42&\textcolor{cyan}{32.01}&\textcolor{cyan}{0.936}&57.12&{34.41}&{0.943}&65.69 &\textcolor{cyan}{34.13} &\textcolor{cyan}{0.944}&35.16 &\textcolor{cyan}{34.08} &\textcolor{cyan}{0.941} &44.82 &{0.987} \\
DeepRFTv3-L &\textcolor{magenta}{34.02}&\textcolor{magenta}{0.953}&\textcolor{magenta}{45.27} &\textcolor{magenta}{32.49}&\textcolor{magenta}{0.941}&\textcolor{magenta}{58.10}&\textcolor{magenta}{35.20}&\textcolor{magenta}{0.951}&\textcolor{magenta}{66.31} &\textcolor{magenta}{34.37} &\textcolor{magenta}{0.947}&\textcolor{magenta}{35.90} &\textcolor{magenta}{34.17} &\textcolor{magenta}{0.942}&\textcolor{magenta}{45.57} &\textcolor{magenta}{0.988} \\
\arrayrulecolor{black}\bottomrule[0.15em]
\end{tabular}
}
\end{center}
\end{table*}
\begin{table*}[t!]
\footnotesize
\begin{center}
\vspace{-1em}
\caption{Comparisons on OOD datasets.}
\label{tab:test2}
\setlength{\tabcolsep}{1pt}
\resizebox{1\linewidth}{!}{
\begin{tabular}{l| cc | cc |cc |cc |cc | cc | cc  }
\toprule[0.15em]
\multicolumn{1}{c|}{} & \multicolumn{2}{c|}{\textbf{MIORe}}& \multicolumn{2}{c|}{\textbf{VAR-MIORe}} &\multicolumn{2}{c|}{\textbf{BSD1}}& \multicolumn{2}{c|}{\textbf{BSD2}}& \multicolumn{2}{c|}{\textbf{BSD3}} &  \multicolumn{2}{c}{\textbf{FoundIR}} &  \textbf{FLOPs} & \textbf{Mem} \\
 \textbf{Method}&{CLIPIQA}&MUSIQ&{CLIPIQA}&MUSIQ &{CLIPIQA}&MUSIQ &{CLIPIQA}&MUSIQ&{CLIPIQA}&MUSIQ & {CLIPIQA}&MUSIQ& (G) & (GB) \\
 \midrule[0.15em]
 GT &0.3076 &55.70 &0.3045 &56.81 &0.2695&55.05 &0.2919&51.42 &0.2658&49.70 &0.3207 &61.42 & - & -\\
Blur &0.2120 &36.75&0.1953 &26.85 &0.2097 &40.35&0.1854 &34.15&0.1740 &30.89 &0.1477&23.76 & - & -\\
\arrayrulecolor{black!30}\midrule
MIMOUNet  &0.2422  &44.41 &0.2147 &32.97 & 0.2415  &49.08  &0.2306  &42.24 &0.2111&41.53 &0.2756  &48.21   & 67 & 12 \\
DeepRFT  &\textcolor{cyan}{0.2797} &\textcolor{cyan}{52.01} &0.2299 &\textcolor{cyan}{39.91} &0.2656 &{52.70} &0.2787 &{48.23} &0.2395 &47.76 &0.3146 &53.90 & 80 & 24\\
NAFNet &0.2704 &51.00&0.2224 &37.88 &0.2886 &53.74&0.2923 &49.28 &0.2649 & 48.02&0.3102 &53.82  & 63 & 35\\
LoFormer &0.2766 &{51.95}&0.2272 &37.17  &0.2708 &55.01 & 0.2615 &49.48 &0.2362 &48.21 &0.3142 &54.49  & 146 & 179\\
PromptIR &0.2636 &49.26 &0.2197 &35.13 &0.2867 &53.97&0.2811 &48.41&0.2492 &42.15 &0.3099 &53.55& 158 & 212\\
MoCEIR&0.2601 &48.69&0.2232 &35.80 &0.2801&54.10&0.2830 &49.35 &0.2513 &47.27&0.3122 &54.48& 87 & 136\\
BioIR  &0.2689 &50.49&0.2265 &37.67 &0.2990&54.82&{0.2943} &{49.13} &0.2747 &\textcolor{magenta}{49.10} &\textcolor{cyan}{0.3239} &{55.83}& 126 & 233\\
EVSSM&0.2624 &48.52&0.2248 &36.46&0.2916&53.41&0.2856 &47.88 &0.2617 &47.93&0.3172&55.62& 122 & 236\\
\arrayrulecolor{black!30}\midrule
DeepRFTv2 &{0.2819} &{52.17}&{0.2297} &38.99&{0.3060}&{55.44}&{0.2903} &{49.30} &{0.2784} &{47.88}&0.3276&{57.09}& 72 & 88 \\
DeepRFTv3-B &{0.2784} &{51.70}&\textcolor{cyan}{0.2317} &38.61&\textcolor{cyan}{0.3006}&\textcolor{magenta}{55.62}&\textcolor{magenta}{0.3031} &\textcolor{magenta}{51.17} &\textcolor{cyan}{0.2749} &\textcolor{cyan}{49.09}&0.3231&\textcolor{cyan}{56.70}& 80 & 103\\
DeepRFTv3-L&\textcolor{magenta}{0.2820} &\textcolor{magenta}{52.23}&\textcolor{magenta}{0.2367} &\textcolor{magenta}{40.44} &\textcolor{magenta}{0.3017}&\textcolor{cyan}{55.11} &\textcolor{cyan}{0.3010} &\textcolor{cyan}{50.08}&\textcolor{magenta}{0.2781} &{48.79} &\textcolor{magenta}{0.3319}&\textcolor{magenta}{57.72}& 130 & 265\\
\arrayrulecolor{black}\bottomrule[0.15em]
\end{tabular}
}
\end{center}
\vspace{-1.em}
\end{table*}

\subsection{Multi-Source Dataset Construction}

To rigorously train and evaluate our unified model, we establish a comprehensive multi-source dataset, categorized into in-domain training and out-of-distribution (OOD).

\noindent\textbf{In-Domain Training Datasets.} We construct our training corpus using six high-quality datasets: GoPro~\cite{nah2017deep}, HIDE~\cite{Shen2019human}, REDS~\cite{Nah2021ntire}, RSBlur~\cite{rim2022RSBlur}, RealBlur~\cite{Rim2020real}, and LOLBlur~\cite{zhou2022lednet}. These datasets feature pristine ground truths (GTs) and cover diverse scenarios (synthetic/real blurs, low-light, and dynamic motions), providing a robust foundation for unified deblurring.

\noindent\textbf{OOD Testing Datasets.} 
To evaluate cross-domain generalization, we select six challenging OOD test sets: BSD1(1ms-8ms) / BSD2(2ms-16ms) / BSD3(3ms-24ms)~\cite{zhong2023BSD}, MIORe / VAR-MIORe~\cite{ciubotariu2025MIORe}, and FoundIR~\cite{li2025foundir}.

\section{Experiment}

\begin{figure*}[t!]
\begin{center}
    \includegraphics[width=1.\linewidth]{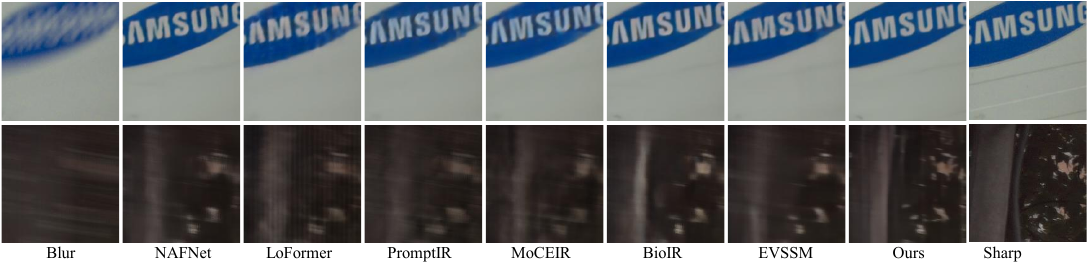}
\end{center}
\vspace{-1.em}
\caption{Visual comparison on RSBlur and VAR-MIORe.}
\label{fig:vis}
\vspace{-0.7em}
\end{figure*}

\subsection{Experimental Setup}
\label{sec:dataset-detail}
\textbf{DeepRFTv3 Architecture.} 
We introduce two variants, DeepRFTv3-B and DeepRFTv3-L. As shown in Fig.~\ref{fig:arch}(a), DeepRFTv3 adopts a 3-scale U-Net architecture, with detailed configurations summarized in Table~\ref{tab:ablation}.

\noindent\textbf{Comparative Methods.} We selected representative end-to-end methods MIMO-UNet~\cite{Cho2021rethinking}, DeepRFT~\cite{XintianMao2023DeepRFT}, NAFNet~\cite{chen2022simple}, LoFormer~\cite{mao2024Loformer}, EVSSM~\cite{kong2025EVSSM}, DeepRFTv2~\cite{xintm2025DeepRFTv2},  PromptIR~\cite{potlapalli2023promptir}, MoCEIR~\cite{zamfir2025MoCEIR} and BioIR~\cite{cui2026BioIR} from both motion deblurring and All-in-One image restoration. We retrain all baselines under the unified training protocol. Beyond evaluating the restoration performance, we also assess the practical efficiency of the models. Specifically, we calculate the computational cost (FLOPs) and the total training memory footprint (Mem) for each configuration, ensuring a comprehensive comparison.

\noindent\textbf{Configuration.} 
All models are trained for 300K iterations using Adam~\cite{kingma2014adam} ($\beta_1=0.9$, $\beta_2=0.999$, weight decay 1$\times$10$^{-2}$), initial learning rate 5$\times$10$^{-4}$. The learning rate is steadily decreased to 1$\times$10$^{-6}$ with the cosine annealing schedule. The batch size is set to 16, and the patch size is $256 \times 256$. To solve the spatial-variant problem, patch-based FCL is used instead of image-based.

\noindent\textbf{Loss Function.} 
All of the models are supervised exclusively by deblurring loss, defined as the sum of the L1 loss ($\ell_{1}$) and the Frequency Reconstruction loss ($\ell_{freq}$): $\ell_{total}=\ell_{1}+0.01\ell_{freq}$.

\noindent\textbf{Evaluation.}  To mitigate evaluation bias from inconsistent GT quality, we adopt a hybrid protocol. For in-domain datasets, we use PSNR and SSIM. For OOD datasets, we rely mainly on robust no-reference metrics (CLIP-IQA~\cite{wang2023CLIPIQA} and MUSIQ~\cite{ke2021musiq}) to assess perceptual quality.

\subsection{Comparisons with Representative Methods}

\noindent\textbf{Results on In-Domain Datasets.} 
As shown in Table~\ref{tab:test1}, recent models (PromptIR, MoCEIR) achieve suboptimal performance (31.94 dB and 32.36 dB on GoPro) on complex motion deblurring. This limitation primarily arises because their spatial feature routing lacks an explicit mechanism to model the convolutional nature of blur. While methods like DeepRFT demonstrate the effectiveness of frequency-domain learning over the spatial baseline MIMO-UNet, they typically impose severe training memory burdens due to dense spatial-frequency transformations. By introducing the Fourier Convolution Learning (FCL) module exclusively at the encoder-decoder transitions, DeepRFTv3-L boosts the GoPro PSNR from the strong baseline EVSSM's 33.26 dB to 34.02 dB. This 0.76 dB gain incurs a small computational overhead of only 8G FLOPs (from 122G to 130G), demonstrating that our FCL effectively models the convolutional degradation of blur.

\noindent\textbf{Results on Out-of-Distribution Datasets.} 
For OOD evaluation (Table~\ref{tab:test2} and Table~\ref{tab:PSNRonOOD} in Appendix), real-world GT quality inevitably degrades as exposure time increases (GT MUSIQ drops from 55.05 to 49.70 on BSD), justifying our reliance on robust no-reference metrics (CLIPIQA and MUSIQ). Equipped with FCL, our DeepRFTv3-L (adopting the identical EVSBlock and depth as EVSSM) delivers marked perceptual improvements across unseen scenarios, notably outperforming EVSSM on MIORe (52.23 vs. 48.52 MUSIQ) and VAR-MIORe (0.2367 vs. 0.2248 CLIPIQA). Furthermore, to validate structural efficiency, DeepRFTv3-B integrates the same BioBlock as BioIR into a shallower backbone. Despite requiring significantly lower computation and memory (80G FLOPs / 103GB Mem vs. BioIR's 126G / 233GB), DeepRFTv3-B surpasses BioIR across nearly all OOD metrics, trailing only marginally on BSD3 MUSIQ (49.09 vs. 49.10) and FoundIR CLIPIQA (0.3231 vs. 0.3239). These results indicate that FCL improves cross-domain performance while introducing limited computational overhead.

\noindent\textbf{Visual Comparisons.} 
As shown in Fig.~\ref{fig:vis} and Fig.~\ref{fig:GoPro} $\sim$ Fig.~\ref{fig:FoundIR} (Appendix), while existing methods often leave noticeable residual blur, DeepRFTv3-L consistently recovers crisp edges and fine textures. Across both in-domain and unseen OOD scenarios, our DeepRFTv3 effectively removes blur artifacts to reconstruct accurate and visually pleasing details.


\begin{table*}[t!]
\footnotesize
\begin{center}
\vspace{-0.5em}
\caption{Ablation for Fourier Convolution Learning. Exp is the index of each experiment. Dim is the channel number of Level 1. Num means the number of basicblock of FCL. Mode means get $\mathcal{F}(\mathrm{x}_{out})$ via $\odot$ (multiplication) or $\oplus$ (addition). MS denotes applying FCL in multi-scale or not. Exp \underline{6} and \underline{13} are trained without auxiliary supervision. Exp $\widetilde{13}$ placed the FCL to the end of each decoder.} 
\label{tab:ablation}
\setlength{\tabcolsep}{0.5pt}
\resizebox{1.\linewidth}{!}{
\begin{tabular}{ c|ccc | ccccc | cccc | cc }
\toprule[0.15em]
& \multicolumn{3}{c|}{\textbf{Backbone}} & \multicolumn{5}{c|}{\textbf{Fourier Convolution Learning}} & \multicolumn{1}{c}{\textbf{GoPro}}& \multicolumn{1}{c}{\textbf{HIDE}} & \multicolumn{1}{c}{\textbf{RSBlur}}& \multicolumn{1}{c}{\textbf{LOLBlur}} & FLOPs &Mem\\

Exp &Basicblock & Level1 $\sim$ 3   & Dim  & Basicblock &mode &MS  & Num  & Dim & PSNR & PSNR & PSNR & PSNR & (G) & (GB)\\
  \midrule[0.15em]
0 & NAFBlock & [1, 4, 8] & 32  & - & - & -  & -  & - & 31.96 & 30.27 & 33.26 & 43.32 & 14.20 &21\\
\arrayrulecolor{black!30}\midrule
1 & NAFBlock & [1, 4, 8]& 32 & ConvBlock  & $\oplus$ & $\times$    &4 & 48 & 32.08 & 30.41 & 33.35 & 43.49 & 15.20 &22\\
2 & NAFBlock & [1, 4, 8]& 32 & Conv$1\times1$  & $\oplus$ & $\times$    &4 & 48 & 32.11 & 30.44 & 33.39 & 43.51 & 15.17 &22\\
3 &  NAFBlock & [1, 4, 8]& 32 & ConvBlock  & $\odot$ & $\times$    &4 & 48 & 32.23 & 30.56 & 33.51 & 43.62 & 15.20 &22\\
4 & NAFBlock & [1, 4, 8] & 32 &  MLPBlock & $\odot$ & $\times$  & 4 & 48 & 32.27 & 30.59 & 33.52 & 43.61 & 15.36 & 22\\

5 & NAFBlock & [1, 4, 8]& 32  & GatedBlock & $\oplus $ & $\times$  &4 & 48 & 32.23 & 30.57 & 33.51 & 43.63 & 15.52 & 22\\
\underline{6} & NAFBlock & [1, 4, 8]& 32  & GatedBlock & $\odot$ & $\times$  &4 & 48 & 32.21 & 30.53 & 33.53 & 43.65 & 15.52 & 22\\
6 & NAFBlock & [1, 4, 8]& 32  & GatedBlock & $\odot$ & $\times$  &4 & 48 & 32.26 & 30.59 & 33.56 & 43.68 & 15.52 & 22\\
7 & NAFBlock & [1, 4, 8]& 32 & SS2DBlock & $\odot$ & $\times$   & 4 & 48 & 32.26 & 30.62 & 33.56 & 43.64 & 16.47 & 24\\
\arrayrulecolor{black!30}\midrule

8 & NAFBlock & [1, 4, 8] & 32  & ConvBlock & $\oplus$ & \checkmark & 4   & 48 & 32.59 & 31.09 & 33.64 & 43.83 & 17.19 & 27\\
9 & NAFBlock & [1, 4, 8] & 32  & ConvBlock & $\odot$& \checkmark & 4   & 48 & 32.74 & 31.23 & 33.75 & 43.94 &  17.19&27 \\
10 & NAFBlock & [1, 4, 8] & 32  & GatedBlock & $\odot$ & \checkmark & 1   & 48 & 32.49 & 31.06 & 33.64 & 43.86 & 16.07 & 25\\
11 & NAFBlock & [1, 4, 8] & 32  & GatedBlock & $\odot$& \checkmark & 4   & 32 & 32.71 & 31.24 & 33.78 & 43.93 & 16.49&27 \\
12 & NAFBlock & [1, 4, 8] & 32  & GatedBlock & $\oplus$ & \checkmark & 4   & 48 & 32.70 & 31.24 & 33.72 & 43.96 &18.31 &28 \\
\underline{13} & NAFBlock & [1, 4, 8] & 32  & GatedBlock & $\odot$ & \checkmark & 4 & 48 & 32.66 & 31.18 & 33.75 & 43.97 &18.31 &28 \\
$\widetilde{13}$ & NAFBlock & [1, 4, 8] & 32  & GatedBlock & $\odot$ & \checkmark & 4   & 48 & 32.62 & 31.05 & 33.70 & 43.81 &18.31 &28 \\
13 & NAFBlock & [1, 4, 8] & 32  & GatedBlock & $\odot$ & \checkmark & 4   & 48 & 32.74 & 31.28 & 33.78 & 44.04 &18.31 &28 \\

14 & NAFBlock & [1, 4, 8] & 32  & SS2DBlock  & $\odot$ &\checkmark & 1   & 48  & 32.56 & 31.08 & 33.65 & 43.88 & 16.79 & 25\\
15 & NAFBlock & [1, 4, 8] & 32  & SS2DBlock & $\odot$ & \checkmark & 4   & 48 & 32.85 & 31.41 & 33.83 & 44.10 & 21.19 &40\\
16 & NAFBlock & [1, 4, 8] & 32  & SS2DBlock & $\odot$ & \checkmark & 8   & 32 & 32.90 & 31.45 & 33.88 & 44.12 & 20.50 &46\\

\arrayrulecolor{black!30}\midrule
17 & BioBlock & [1, 4, 12] & 48 & - & - & -  & -& - & 32.52 & 31.14 & 33.57 & 44.28 & 75 & 95\\
18 & BioBlock & [1, 4, 12] & 48  & GatedBlock & $\odot$& $\times$ & 4   & 48 & 33.24 & 31.72 & 34.02 & 44.78 &77  &96\\
\rowcolor{gray!15}19 & DeepRFTv3-B & [1, 4, 12] & 48  & GatedBlock & $\odot$& \checkmark  & 4 & 48 & 33.48 & 32.01 & 34.13 & 44.82 & 80 & 103\\
\arrayrulecolor{black!30}\midrule
20 &EVSBlock & [6, 6, 12] & 48 & - & - & -  & -& - & 33.26 & 31.70 & 33.91 & 44.94 & 122 & 236\\
21 &EVSBlock & [6, 6, 12] & 48  & SS2DBlock & $\odot$ & $\times$  & 4 & 48 & 33.73 & 32.20  & 34.26 & 45.42 & 125 &248\\
\rowcolor{gray!15}22 &DeepRFTv3-L & [6, 6, 12] & 48  & SS2DBlock & $\odot$ & \checkmark  & 4 & 48 & 34.02 & 32.49  & 34.37 & 45.57 & 130 &265\\
\arrayrulecolor{black}\bottomrule[0.15em]
\end{tabular}
}
\end{center}
\vspace{-1.5em}
\end{table*}

\subsection{Analysis and Discussion}




To fully understand the contribution of each component in our proposed method, we conduct extensive ablation studies and visualization analysis. As summarized in Table~\ref{tab:ablation}, we systematically evaluate the effectiveness of Fourier Convolution Learning, the impact of multi-scale integration, the choice of frequency-domain fusion operators, the basicblock for backbone (NAFBlock~\cite{chen2022simple}, BioBlock~\cite{cui2026BioIR}, EVSBlock~\cite{kong2025EVSSM}), and the basicblock for FCL (ConvBlock Fig.~\ref{fig:arch}(c), MLPBlock Fig.~\ref{fig:arch}(d), GatedBlock Fig.~\ref{fig:arch}(e), SS2DBlock Fig.~\ref{fig:arch}(f)). 

\noindent\textbf{Effectiveness of Fourier Convolution Learning.} 
We first evaluate the fundamental impact of integrating FCL into the baseline network. As shown in Table~\ref{tab:ablation}, the baseline model (Exp 0) without FCL achieves 31.96 dB on the GoPro dataset with 14.20G FLOPs and 21GB of memory. By introducing a single-scale FCL module at the encoder-decoder transition, regardless of the lightweight non-linear basicblocks used (ConvBlock in Exp 3, MLPBlock in Exp 4, GatedBlock in Exp 6, or SS2DBlock in Exp 7), the network obtains significant performance gains across all general motion deblurring datasets. In particular, Exp 6 improves the GoPro PSNR to 32.26 dB with only a 1.32G FLOPs and 1GB memory increase. This demonstrates that the FCL module can efficiently empower the model to understand the convolutional degradation process with minimal computational overhead.

\begin{figure*}[t]
\begin{center}
    \includegraphics[width=1.\linewidth]{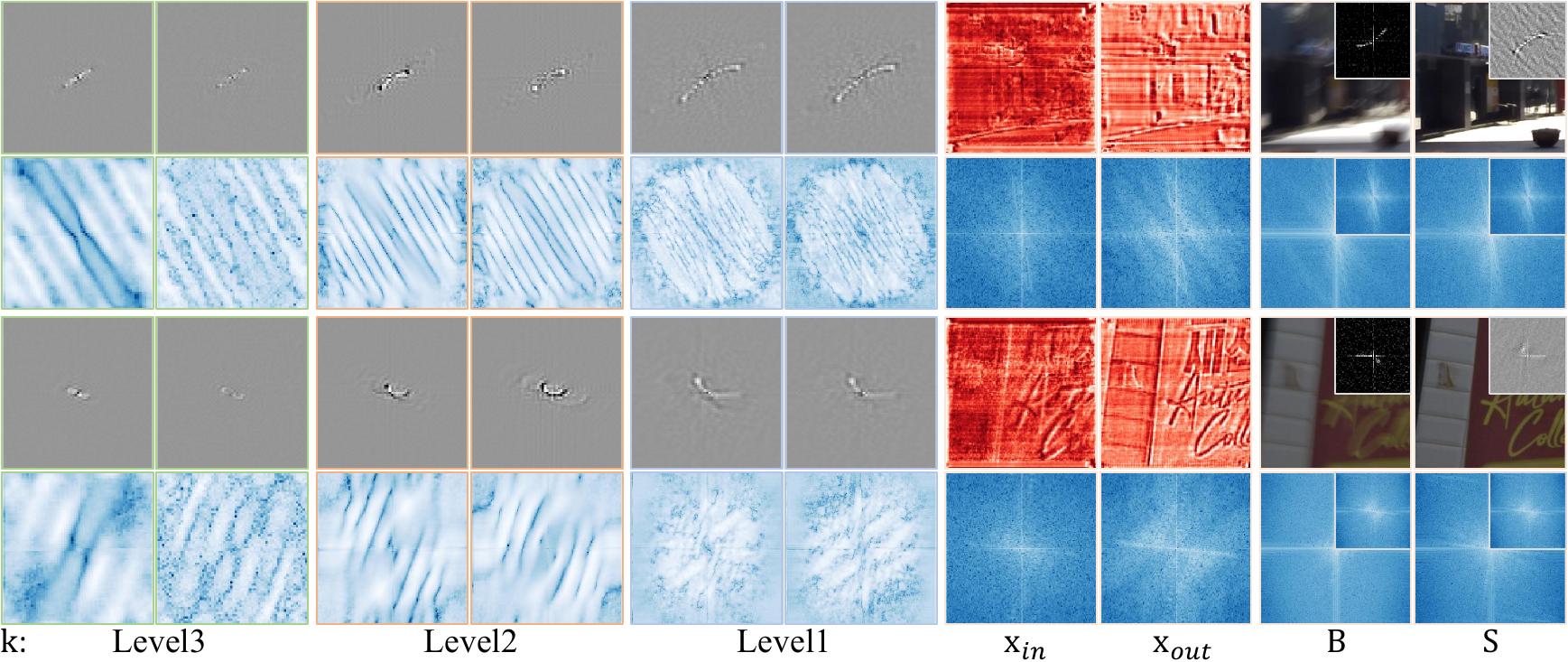}
\end{center}
\vspace{-1.em}
\caption{Visualization of the frequency filters $\mathcal{F}(\mathrm{k})$ and corresponding spatial kernels $\mathrm{k}$ at Levels 1 $\sim$ 3, together with the averaged $\mathrm{x}_{in}~/~\mathrm{x}_{out}$ and the corresponding $\mathrm{B}~/~\mathrm{S}~/~\mathrm{k}_{blur}~/~\mathrm{k}_{deblur}$ on GoPro and RSBlur.}
\label{fig:vis_fcl}
\end{figure*}

\noindent\textbf{Effectiveness of Multi-Scale FCL.} 
Building on the success of single-scale, we further extend FCL to multi-scale (MS) FCL in Table~\ref{tab:ablation}. Comparing Exp 6 (single-scale) with Exp 13 (multi-scale) reveals a comprehensive performance improvement (GoPro PSNR increases from 32.26 dB to 32.74 dB), demonstrating the benefit of multi-scale FCL. Furthermore, we investigate the depth (Num) and width (Dim) of the non-linear basicblocks within the FCL module (Exp 10, 11, 13 and Exp 14, 15, 16). The results indicate that deeper and wider blocks generally yield better gains. For example, increasing the number of SS2DBlocks from 1 (Exp 14) to 4 (Exp 15) improves the GoPro PSNR from 32.56 dB to 32.85 dB. However, this comes with a severe memory trade-off: pushing the depth to 8 (Exp 16) brings only small gains (32.90 dB) but increases the memory to 46GB. 

\noindent\textbf{Convolution \emph{v.s.} Addition.} 
We compare feature integration in Fourier space via addition ($\oplus$) and multiplication ($\odot$) in Table~\ref{tab:ablation}. Across all single- and multi-scale settings (Exp 1 vs. Exp 3, Exp 8 vs. Exp 9), $\odot$ consistently outperforms $\oplus$. This is consistent with our formulation: multiplication in the frequency domain corresponds to convolution in the spatial domain, explicitly encouraging the network to learn the blur degradation. Using $\odot$ for the final integration also performs better when the backbone contains multiplicative gating (GatedBlock).

\noindent\textbf{Auxiliary Supervision or Not.} 
Following existing multi-scale designs (MIMO-UNet~\cite{Cho2021rethinking}, DeepRFT~\cite{XintianMao2023DeepRFT}, ConvIR~\cite{cui2024ConvIR}), we project the intermediate FCL output feature into the image space via simple convolutional layers, which are supervised by downsampled ground-truth images. These projection layers are completely discarded during inference, ensuring no additional computational overhead. As shown in Table~\ref{tab:ablation}, although the performance gains from this supervision are marginal in Exp 6 and Exp 13, we retain this cost-free auxiliary multi-scale guidance in our final framework.

\noindent\textbf{Local Feature Extraction.} As shown in Table~\ref{tab:ablation}, lacking local feature extraction hinders frequency-domain filter learning. Compared to the baseline (Exp 0), simply adding spatial (ConvBlock) or point-wise (Conv$1\times1$) features (Exp 1 and 2) yields marginal gains. However, combining local feature extraction boosts the GoPro PSNR to 32.23 dB (Exp 3), demonstrating that effective filters inherently rely on robust local spatial extraction.

\noindent\textbf{Placement of FCL.} Exp 13 and $\widetilde{13}$ indicate that placing FCL between the encoder and decoder is optimal. Shifting FCL to the end of each decoder like DeepRFTv2~\cite{xintm2025DeepRFTv2} (Exp $\widetilde{13}$) drops performance on both the GoPro (from 32.74 dB to 32.62 dB) and HIDE (from 31.28 dB to 31.05 dB) datasets. Thus, we position FCL in the bottleneck to maximize its image restoration capabilities.

\noindent\textbf{Universality of Fourier Convolution Learning.} 
To evaluate whether FCL generalizes across different backbone architectures, we further evaluate its effectiveness on other distinct backbones (Exp 17-22). As shown in Table~\ref{tab:ablation}, beyond the lightweight NAFBlock, FCL exhibits strong generalization on both the BioBlock~\cite{cui2026BioIR} and the EVSBlock~\cite{kong2025EVSSM}. For instance, the baseline BioBlock model (Exp 17) achieves 32.52 dB in PSNR on the GoPro dataset with 75G FLOPs. Integrating single-scale (Exp 18) and multi-scale FCL (Exp 19, our DeepRFTv3-B) significantly boosts the PSNR to 33.24 dB and 33.48 dB, respectively, while incurring a small computational overhead of only 2G to 5G FLOPs. This consistent scaling trend is similarly observed on the more advanced and heavier EVSBlock backbone ((Exp 20-22). These results indicate that FCL can be integrated into different backbone architectures with limited additional computation.


\noindent\textbf{Visualization of Fourier Convolution Learning.} 
To gain deeper insights into our proposed Fourier Convolution Learning, we visualize the global filters ($\mathcal{F}(\mathrm{k})$) and their spatial convolution kernels ($\mathrm{k}$) in different U-Net scales (Level 1 $\sim$ 3) alongside the corresponding feature transformations ($\mathrm{x}_{in}$ to $\mathrm{x}_{out}$) in Fig.~\ref{fig:vis_fcl} and Fig.~\ref{fig:vis_fcl2} (Appendix). The learned filters show a clear relationship with the underlying blur degradation. As observed in the spatial domain (obtained via IFFT), the learned multi-scale kernels are similar to the theoretical deconvolution kernels ($\mathrm{k}_{deblur}$). In the frequency domain, they adaptively capture the directional frequency responses corresponding to specific motion blurs. Furthermore, by comparing the frequency spectrums of the averaged $\mathrm{x}_{in}$ and $\mathrm{x}_{out}$, it is evident that the element-wise multiplication with these blur-adaptive filters achieves an effective full-frequency enhancement. This visualization provides evidence that our FCL successfully executes input-dependent convolution learning.

\section{Conclusion}
In this paper, we proposed DeepRFTv3, a unified framework for motion deblurring based on Fourier Convolution Learning (FCL). FCL performs blur-adaptive filtering in the Fourier space, allowing the network to explicitly model the convolutional nature of motion blur. We also constructed a multi-source training and evaluation protocol. Experiments on in-domain and OOD datasets show that FCL improves deblurring performance and generalization across different blur sources, with moderate computational overhead. These results demonstrate the effectiveness of frequency filtering learning for unified motion deblurring.
\section{Acknowledgements}
This work was supported by the National Natural Science Foundation of China (Grant No. 62471182, 62475072, 42527804), Fundamental Research Funds for the Central Universities, Shanghai Rising-Star Program (Grant No. 24QA2702100), the Science and Technology Commission of Shanghai Municipality (Grant No. 22S31905800, 25JD1401300, 25xtcx00600, 22DZ2229004)
\section*{LLM Usage Statement}
LLM assisted with polishing only. Authors verified all results.

{\small
\bibliographystyle{ieeenat_fullname}
\bibliography{references}
}
\clearpage

\clearpage
\setcounter{page}{1}

\twocolumn[{%
\renewcommand\twocolumn[1][]{#1}%
\maketitlesupplementary
\begin{center}
    \centering
    \captionsetup{type=figure}
    \vspace{-1.em}    \includegraphics[width=0.92\linewidth]{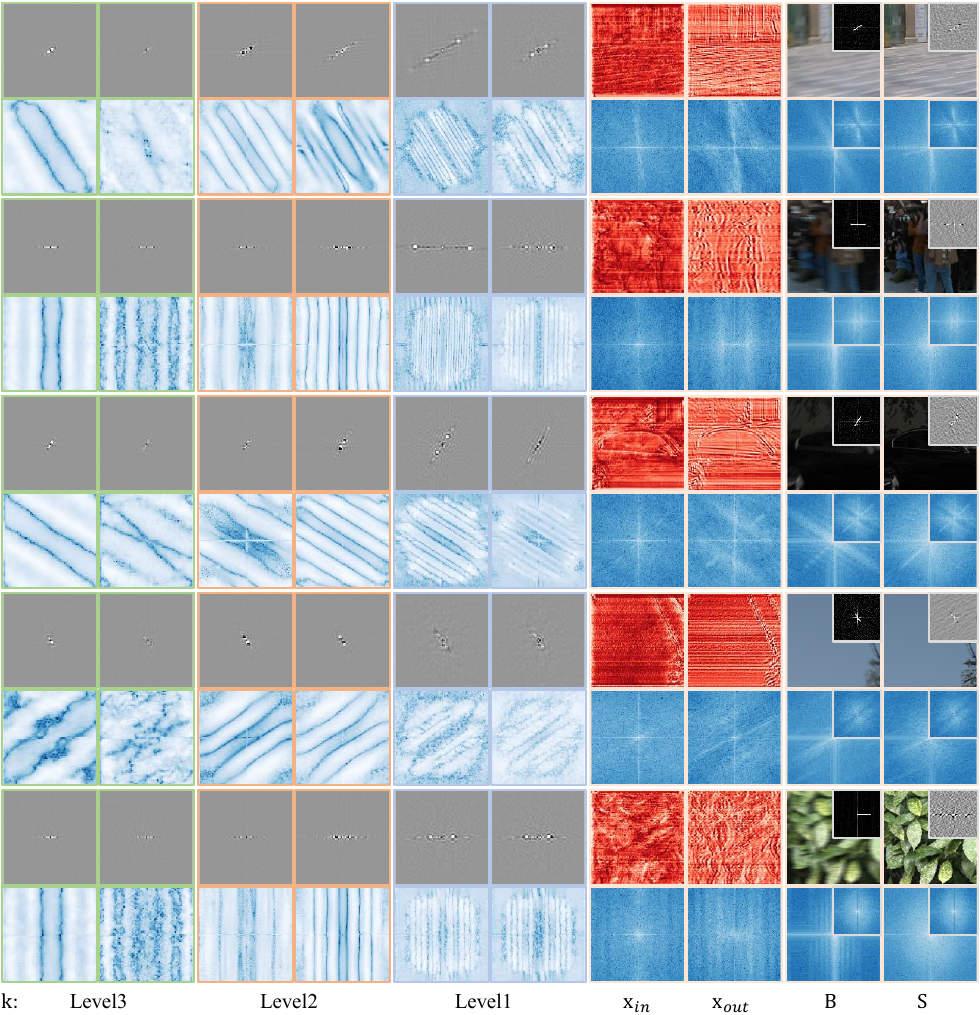}
    \vspace{-0.5em}
    \caption{
    \label{fig:vis_fcl2}
    Visualization of the frequency filters $\mathcal{F}(\mathrm{k})$ and corresponding spatial kernels $\mathrm{k}$ at Levels 1 $\sim$ 3, together with the averaged $\mathrm{x}_{in}~/~\mathrm{x}_{out}$ and the corresponding $\mathrm{B}~/~\mathrm{S}~/~\mathrm{k}_{blur}~/~\mathrm{k}_{deblur}$ on HIDE, REDS, LOLBlur, BSD and FoundIR.} 
\end{center}%
}]

\section{Appendix}

\begin{figure*}[t]
\begin{center}
    \includegraphics[width=1.\linewidth]{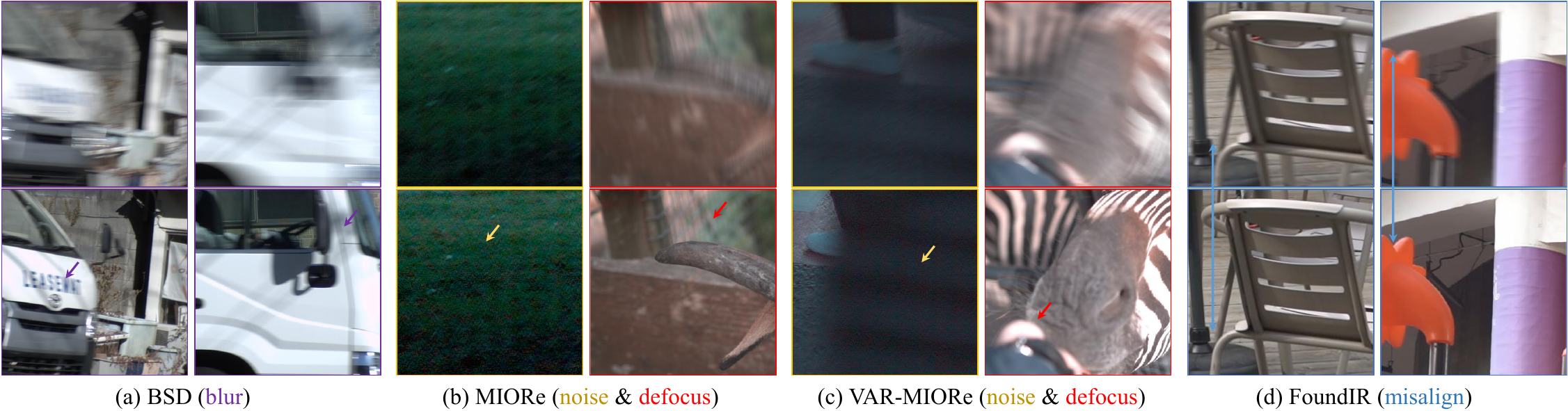}
\end{center}
\vspace{-1.3em}
\caption{Various flaws in OOD datasets.}
\label{fig:err}
\end{figure*}


\begin{table*}[t!]
\centering

\caption{Summary of training and testing datasets.}
\label{tab:train_test}
\resizebox{1.\textwidth}{!}{
\begin{tabular}{l|ccc|cccc}
\toprule
\multicolumn{1}{c|}{} & \multicolumn{3}{c|}{Train}  & \multicolumn{4}{c}{Test} \\
Dataset & Images & Patches & Resolution & Images &Removed &Resolution & Step \\
\midrule
GoPro~\cite{nah2017deep} & 2,103 & 16,824 & 512 $\times$ 512  & 1,111 &0 & 1280 $\times$ 720 & 1 \\
HIDE~\cite{Shen2019human} & 6,397 & 12,794 & 720 $\times$ 720  & 203 &0 & 1280 $\times$ 720 & 10  \\
REDS~\cite{Nah2021ntire} & 24,000 & 48,000 & 720 $\times$ 720  & 299 &1 & 1280 $\times$ 720 & 10 \\
RSBlur~\cite{rim2022RSBlur} & 8,878  & 53,268 & 720 $\times$ 720 & 336 &0 & $\approx$1915 $\times$ 1200 & 10\\
RealBlur~\cite{Rim2020real} & 7,516 & 15,032 & $\approx$660 $\times$ 750  & 1,960 &0 & 640 $\times$ 640 & 1 \\
LOLBlur~\cite{zhou2022lednet} & 10,200 & 20,400 &  640 $\times$ 640  & 600 &0 & 1120 $\times$ 640  &  3 \\
\arrayrulecolor{black!30}\midrule
BSD1(1ms-8ms)~\cite{zhong2023BSD} & - & - & -  & 500 &0 & 640 $\times$ 480 & 6 \\
BSD2(2ms-16ms)~\cite{zhong2023BSD} & - & - & -  & 471 &29 & 640 $\times$ 480 & 6 \\
BSD3(3ms-24ms)~\cite{zhong2023BSD} & - & - & -  & 401 &99& 640 $\times$ 480 & 6 \\
MIORe~\cite{ciubotariu2025MIORe} & - & - & -  & 458 &328 & 1920 $\times$ 1080 &  10 \\
VAR-MIORe~\cite{ciubotariu2025MIORe} & - & - & -  & 203&230 & 1920 $\times$ 1080 & 3 \\
FoundIR~\cite{li2025foundir} & - & - & - & 150  &0& 1920 $\times$ 1080 & 1 \\
\arrayrulecolor{black}\bottomrule
\end{tabular}
}
\end{table*}

\subsection{Visualization of Fourier Convolution Learning}
\noindent\textbf{Extended FCL Visualizations and Dataset Analysis.} 
In Fig.~\ref{fig:vis_fcl2}, we provide additional visualizations of the FCL's input/output features ($x_{in}, x_{out}$) and the learned kernels ($k$) across different hierarchical levels (Level 1 $\sim$ 3) on both in-domain and OOD test sets (including HIDE, REDS, LOLBlur, BSD, and FoundIR). These visual results clearly demonstrate that our FCL module effectively learns blur-adaptive global filters that reflect the distinct motion blur degradation. Furthermore, we visualize the theoretical blur kernel ($k_{blur}$) in the top-right corner of the blur images (column B). Notably, in the FoundIR example (bottom row), the theoretical kernel displays a distinct off-center horizontal shift. This observation visually confirms a severe left-right spatial misalignment between the blur inputs and the ground truths in the FoundIR dataset, directly corroborating the data quality concerns discussed in the main manuscript.

\subsection{Datasets}

\noindent\textbf{OOD Datasets.} As shown in Fig.~\ref{fig:err}, these datasets inherently suffer from acquisition flaws, such as mixed degradations in MIORe, residual GT blur in BSD, and pixel misalignments in FoundIR. To ensure reliable and efficient evaluation, we systematically prune the data (Table~\ref{tab:train_test}). Specifically, we subsample the test sets to a manageable scale and strictly filter out flawed GT-blur pairs. Additionally, high-resolution training images are cropped into patches for better I/O.


\begin{table*}[t!]
\footnotesize
\begin{center}
\caption{Comparisons on ReLoBlur and RealLOL.}
\label{tab:ReLoBlur}
\setlength{\tabcolsep}{1pt}
\resizebox{1.\linewidth}{!}{
\begin{tabular}{ c|cc | cc | cc | cc| cc| cc }
\toprule[0.15em]
& \multicolumn{2}{c|}{\textbf{GT}} & \multicolumn{2}{c|}{\textbf{Blur}} & \multicolumn{2}{c|}{\textbf{PromptIR}} & \multicolumn{2}{c|}{\textbf{MoCEIR}} & \multicolumn{2}{c|}{\textbf{EVSSM}}& \multicolumn{2}{c}{\textbf{DeepRFTv3-L}} \\
Dataset&{CLIPIQA}&MUSIQ&{CLIPIQA}&MUSIQ&{CLIPIQA}&MUSIQ&{CLIPIQA}&MUSIQ&{CLIPIQA}&MUSIQ &{CLIPIQA}&MUSIQ\\
 
  \midrule[0.15em]
ReLoBlur& 0.2103&40.14& 0.1889&37.20 &0.2546&41.30&0.2169&39.85 &0.2841&43.61 &0.2913&43.74\\
\arrayrulecolor{black!30}\midrule

RealLOL&- &- &0.1370 &26.12 &0.1979&29.74&0.2028&29.69 &0.1995&30.04 &0.2088&30.28\\

\arrayrulecolor{black}\bottomrule[0.15em]
\end{tabular}
}
\end{center}
\end{table*}
\begin{figure*}[t]
\begin{center}
    \includegraphics[width=1.\linewidth]{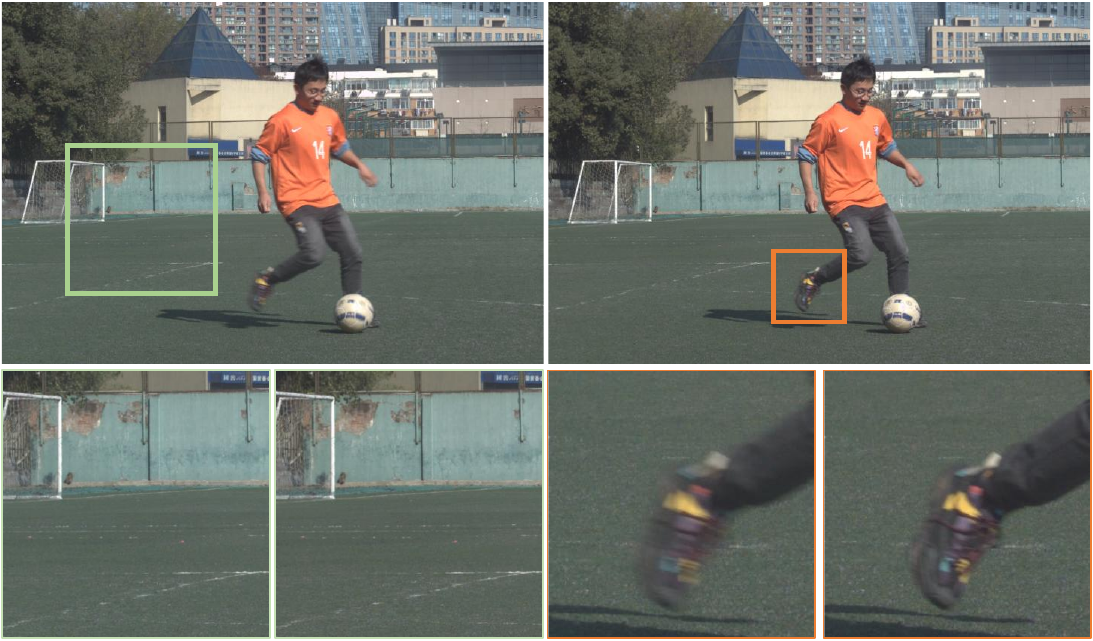}
\end{center}
\vspace{-1.3em}
\caption{Color shift and residual blur of ReLoBlur dataset.}
\label{fig:err_ReLoBlur}
\end{figure*}

\noindent\textbf{ReLoBlur and RealLOL.} 
As shown in Table~\ref{tab:ReLoBlur} (provided only as diagnostic references and not used to support the quantitative benchmark claim), our proposed DeepRFTv3-L demonstrates promising cross-domain generalization on both the ReLoBlur~\cite{li2023ReLoBlur} and RealLOL~\cite{zhou2022lednet} datasets. Despite these superior results, we deliberately exclude both datasets from our main experimental comparisons due to their inherent data defects, which could severely compromise evaluation integrity.

Specifically, the ReLoBlur dataset, captured via a dual-camera beam-splitting system, exhibits noticeable color shifts between the blur inputs and the ground truths (GTs). More severely, the majority of its GT images still contain perceptible motion blur artifacts (shown in Fig.~\ref{fig:err_ReLoBlur}). This flaw is quantitatively evidenced in Table~\ref{tab:ReLoBlur}, where the no-reference scores of the original blurry inputs (CLIPIQA: 0.1889, MUSIQ: 37.20) and the so-called GT images (CLIPIQA: 0.2103, MUSIQ: 40.14) are exceptionally close. Comparing restoration models on such a dataset is unreliable. 

Similarly, the RealLOL dataset lacks corresponding GT images entirely. Furthermore, it consists predominantly of extreme low-light scenarios. Since no-reference perceptual metrics (CLIPIQA and MUSIQ) are sensitive to global image brightness and contrast, applying them to such dark scenes might inaccurately reflect illumination variations rather than the actual deblurring quality. So we ultimately omitted RealLOL from our core benchmark comparisons.

\begin{table*}[t!]
\begin{center}
\caption{Performance of NAFNet32 trained by different \colorbox{color4}{datasets}.}
\label{tab:dataset_ablation}
\setlength{\tabcolsep}{5pt}
\resizebox{0.8\linewidth}{!}{
\begin{tabular}{ l|cccc|cc }
\toprule[0.15em]
&GoPro & HIDE & RealBlur& LOLBlur & REDS & RSBlur\\
\arrayrulecolor{black!30}\midrule
train&PSNR&PSNR&PSNR&PSNR&PSNR&PSNR\\
 \arrayrulecolor{black}\midrule
GoPro&\colorbox{color4}{32.60} & 30.44 & 30.04 & 35.24 &27.08 & 26.22\\
GoPro+HIDE&\colorbox{color4}{32.88} & \colorbox{color4}{31.11} & 29.99 & 34.94 &27.15 & 28.62\\
GoPro+HIDE+RealBlur&\colorbox{color4}{32.53} & \colorbox{color4}{30.80} & \colorbox{color4}{33.67} & 37.60 &27.81 & 30.10\\
GoPro+HIDE+LOLBlur&\colorbox{color4}{32.63} & \colorbox{color4}{30.93} & 30.63 & \colorbox{color4}{43.46} &28.58 & 28.68\\
GoPro+HIDE+RealBlur+LOLBlur&\colorbox{color4}{32.45} & \colorbox{color4}{30.75} & \colorbox{color4}{33.72} & \colorbox{color4}{43.28} &29.62 & 30.31\\
\arrayrulecolor{black}\bottomrule[0.15em]
\end{tabular}
}
\end{center}
\end{table*}

\begin{table*}[t!]
\footnotesize
\begin{center}
\caption{Quantitative comparisons on OOD. BSD-Avg is the average of BSD1, BSD2 and BSD3.}
\label{tab:PSNRonOOD}
\setlength{\tabcolsep}{2pt}
\resizebox{1.\linewidth}{!}{
\begin{tabular}{ c|cc | cc | cc | cc| cc | cc| cc }
\toprule[0.15em]
& \multicolumn{2}{c|}{\textbf{GT}} & \multicolumn{2}{c|}{\textbf{Blur}} & \multicolumn{2}{c|}{\textbf{PromptIR}} & \multicolumn{2}{c|}{\textbf{MoCEIR}} & \multicolumn{2}{c|}{\textbf{EVSSM}}& \multicolumn{2}{c}{\textbf{DeepRFTv3-B}} & \multicolumn{2}{c}{\textbf{DeepRFTv3-L}} \\
Dataset&{PSNR}&SSIM&{PSNR}&SSIM&{PSNR}&SSIM&{PSNR}&SSIM&{PSNR}&SSIM &{PSNR}&SSIM &{PSNR}&SSIM\\
 
  \midrule[0.15em]
BSD-Avg& -&1.000&27.62&0.859&30.41&0.934&30.72&0.936&30.80&0.940
&30.49&0.940&30.60&0.940\\
\arrayrulecolor{black!30}\midrule
MIORe&- &1.000 &30.36 &0.824 &31.94&0.874&31.75&0.872 &31.85&0.874
&32.50&0.887 &31.12&0.884\\
\arrayrulecolor{black!30}\midrule
VAR-MIORe& - &1.000 &24.38 &0.690 &26.33&0.749&26.23&0.748 &26.64&0.754 &27.12&0.769&27.04&0.773\\

\arrayrulecolor{black}\bottomrule[0.15em]
\end{tabular}
}
\end{center}
\end{table*}
\subsection{Impact of Multi-Source Training Dataset} 
To further investigate the impact of training on combined datasets, we evaluate NAFNet32~\cite{chen2022simple} using different training data mixtures, as shown in Table~\ref{tab:dataset_ablation}. When combining datasets that share similar camera devices and generation pipelines (GoPro and HIDE), the model effectively aggregates homogeneous information, leading to mutual enhancement on these specific in-domain datasets. For instance, adding HIDE to the training set boosts the PSNR on GoPro from 32.60 dB to 32.88 dB, and on HIDE from 30.44 dB to 31.11 dB. However, this combination provides limited improvement on unseen scenarios. 

Conversely, as we introduce more diverse, multi-source datasets (such as RealBlur and LOLBlur) into the training mixture, a characteristic trade-off emerges. The inclusion of heterogeneous blur patterns prevents the network from overfitting to a specific dataset bias, which results in a slight performance degradation on individual in-domain test sets (in the fully combined setting, the PSNR on GoPro and HIDE drops to 32.45 dB and 30.75 dB, respectively). Crucially, however, this diversification significantly enhances the model's cross-domain generalization capability. As shown in the rightmost columns of Table~\ref{tab:dataset_ablation}, the OOD performance steadily increases with the addition of diverse datasets, ultimately achieving the highest PSNR scores of 29.62 dB on REDS and 30.31 dB on RSBlur. This justifies our motivation for constructing a unified, multi-source training framework to tackle real-world motion blurs.

\subsection{Quantitative Comparisons on OOD}
The quantitative evaluation on out-of-distribution (OOD) datasets in Table~\ref{tab:PSNRonOOD} demonstrates that the DeepRFTv3 series exhibits exceptional cross-domain generalization and robustness when facing unseen degradation scenarios. Although EVSSM achieves the highest PSNR on the BSD-Avg dataset, both models in the DeepRFTv3 series obtain the joint-best SSIM of 0.940. On the more challenging MIORe and VAR-MIORe datasets, the DeepRFTv3 models establish a significant advantage, comprehensively outperforming comparative methods such as PromptIR, MoCEIR, and EVSSM. Specifically, DeepRFTv3-B achieves the highest PSNR (32.50 dB) and SSIM (0.887) on MIORe, and also secures the highest PSNR (27.12 dB) on VAR-MIORe; meanwhile, DeepRFTv3-L obtains the optimal SSIM (0.773) on the VAR-MIORe dataset.

\subsection{Visualization Comparisons}
\begin{figure*}[t]
\begin{center}
    \includegraphics[width=1.\linewidth]{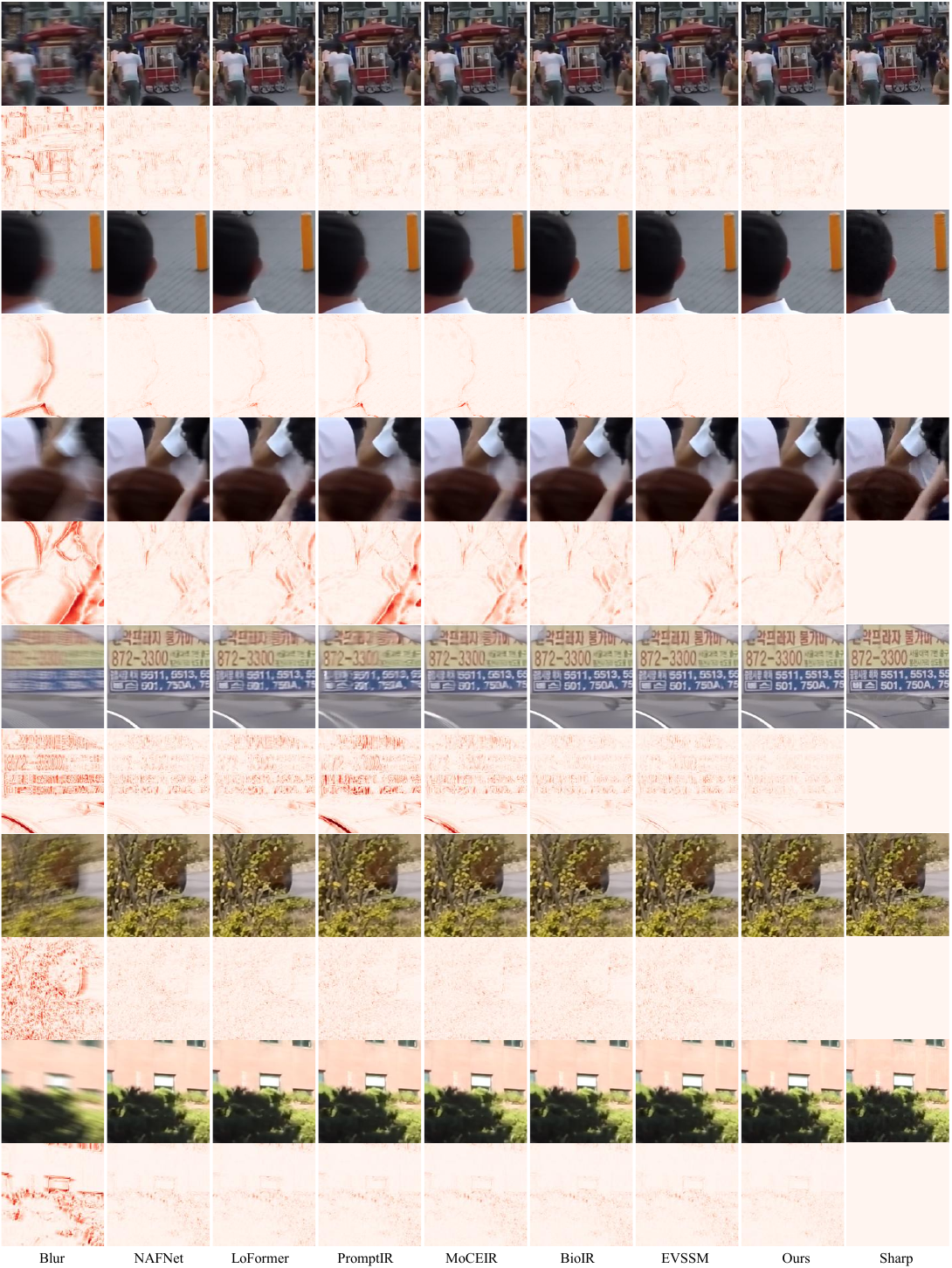}
\end{center}
\vspace{-1.3em}
\caption{Visualization comparisons on GoPro dataset.}
\label{fig:GoPro}
\vspace{-1.em}
\end{figure*}

\begin{figure*}[t]
\begin{center}
    \includegraphics[width=1.\linewidth]{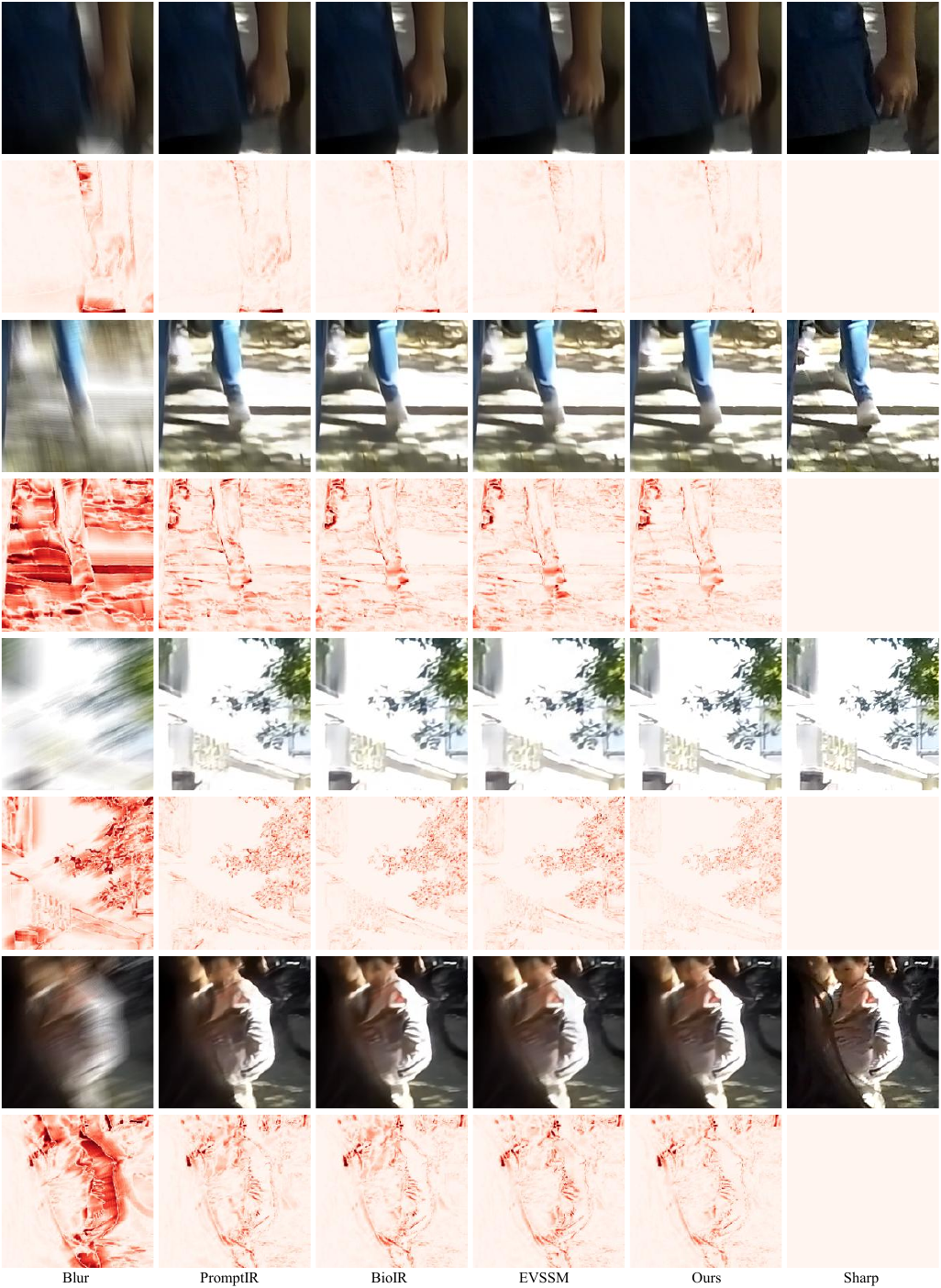}
\end{center}
\vspace{-1.3em}
\caption{Visualization comparisons on HIDE dataset.}
\label{fig:HIDE}
\vspace{-1.em}
\end{figure*}

\begin{figure*}[t]
\begin{center}
    \includegraphics[width=1.\linewidth]{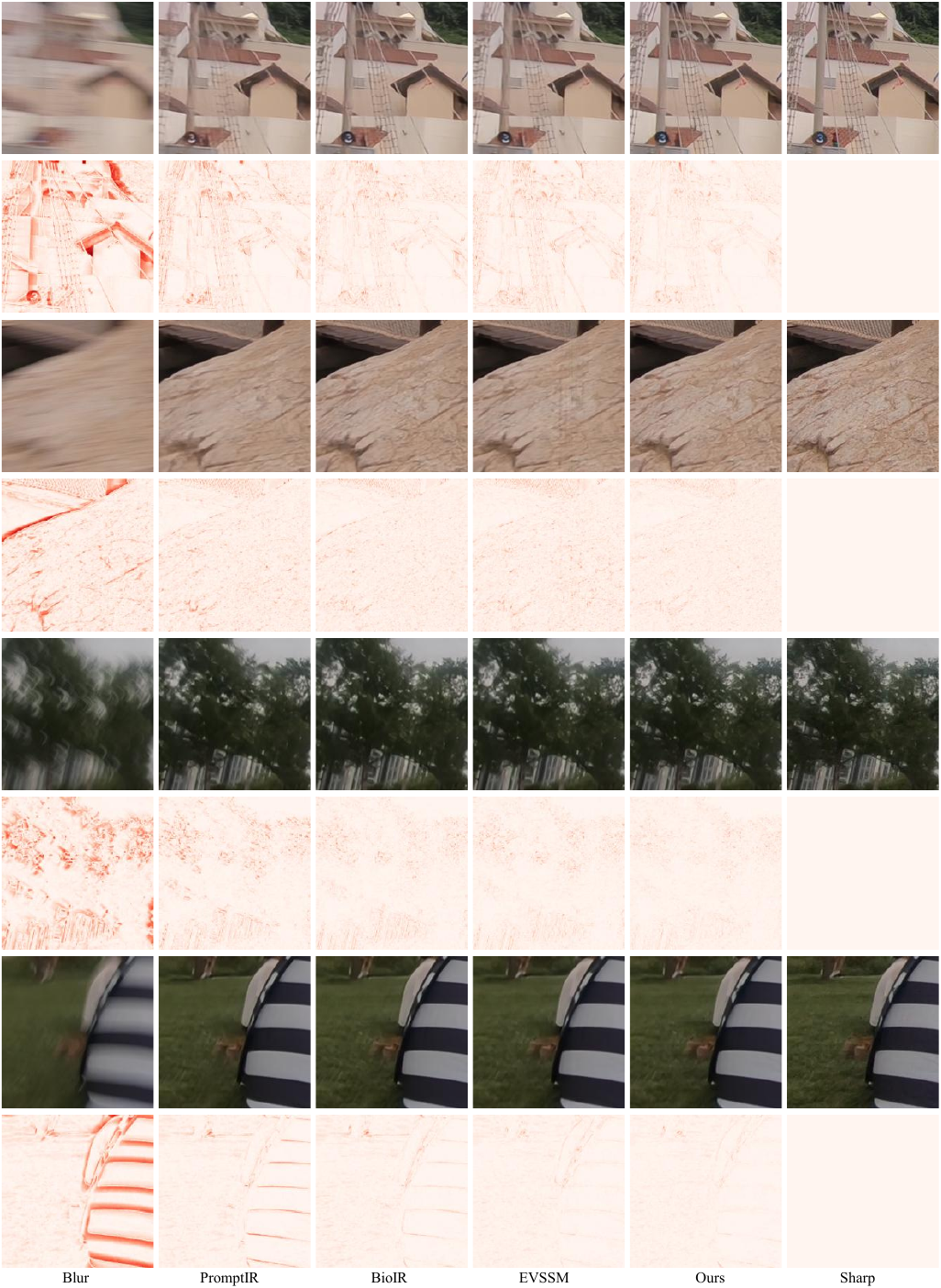}
\end{center}
\vspace{-1.3em}
\caption{Visualization comparisons on REDS dataset.}
\label{fig:REDS}
\vspace{-1.em}
\end{figure*}

\begin{figure*}[t]
\begin{center}
    \includegraphics[width=1.\linewidth]{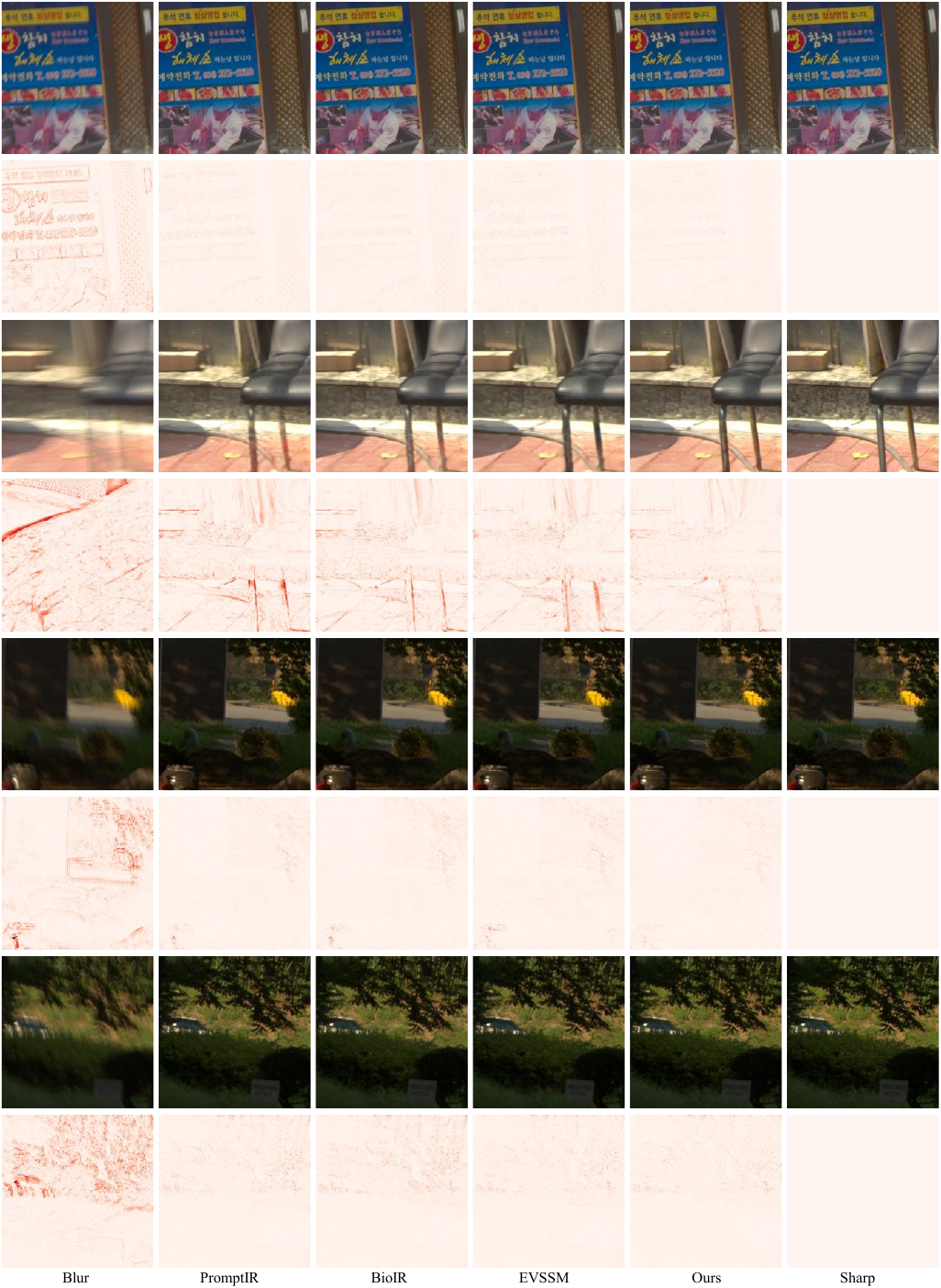}
\end{center}
\vspace{-1.3em}
\caption{Visualization comparisons on RSBlur dataset.}
\label{fig:RSBlur}
\vspace{-1.em}
\end{figure*}

\begin{figure*}[t]
\begin{center}
    \includegraphics[width=1.\linewidth]{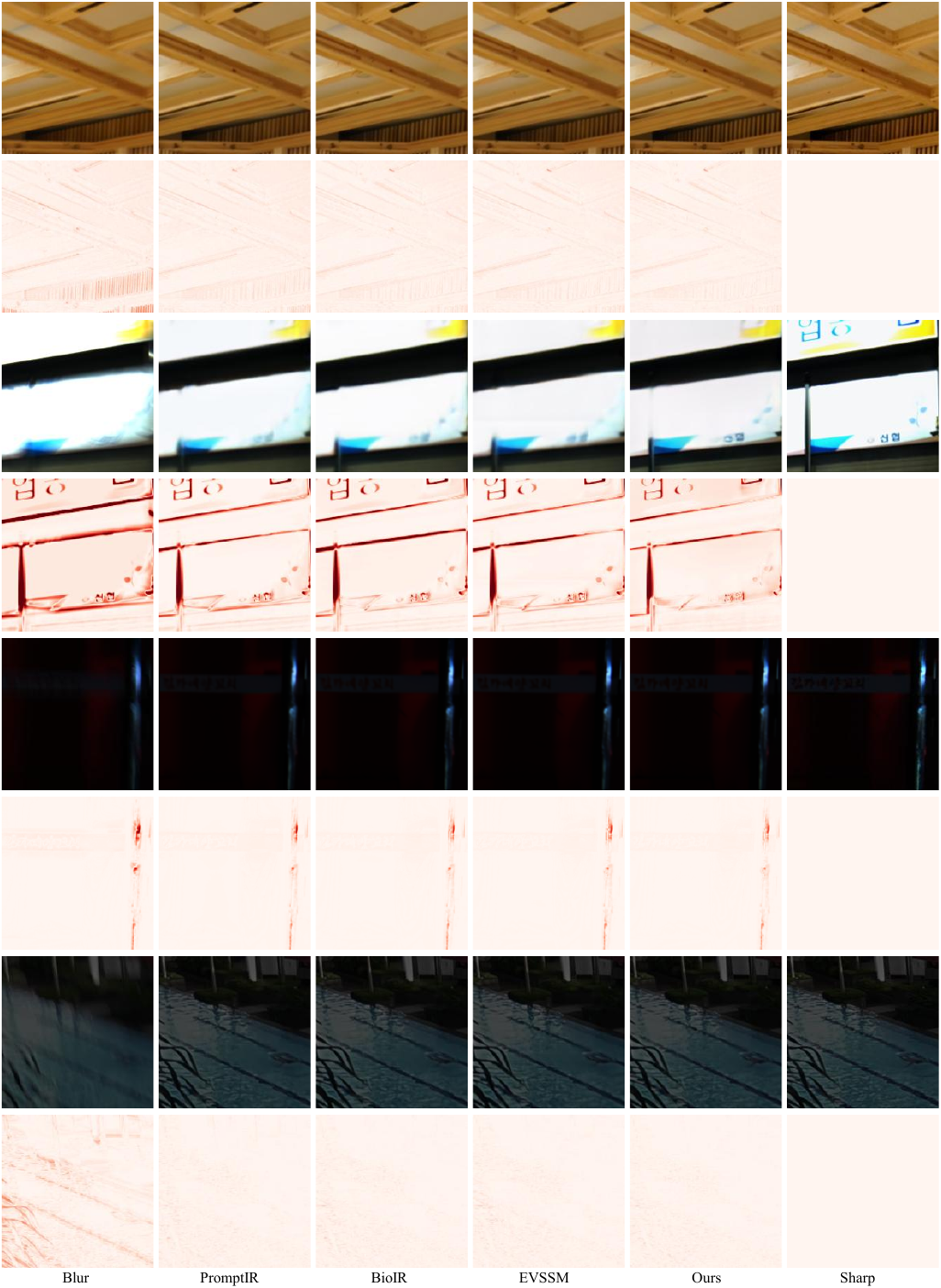}
\end{center}
\vspace{-1.3em}
\caption{Visualization comparisons on RealBlur and LOLBlur dataset.}
\label{fig:lowlight}
\vspace{-1.em}
\end{figure*}

As shown in Fig.~\ref{fig:GoPro} $\sim$ Fig.~\ref{fig:lowlight}, we provide detailed visual comparisons across multiple in-domain datasets, including GoPro, HIDE, REDS, RSBlur, RealBlur, and LOLBlur. For instance, on the GoPro (Fig.~\ref{fig:GoPro}) and RSBlur (Fig.~\ref{fig:RSBlur}) datasets, DeepRFTv3-L restores legible sign numbers and poster letters. On the HIDE (Fig.~\ref{fig:HIDE}) and REDS (Fig.~\ref{fig:REDS}) datasets, the leg contours of pedestrians and the textures of striped clothing are also well recovered.

\begin{figure*}[t]
\begin{center}
    \includegraphics[width=1.\linewidth]{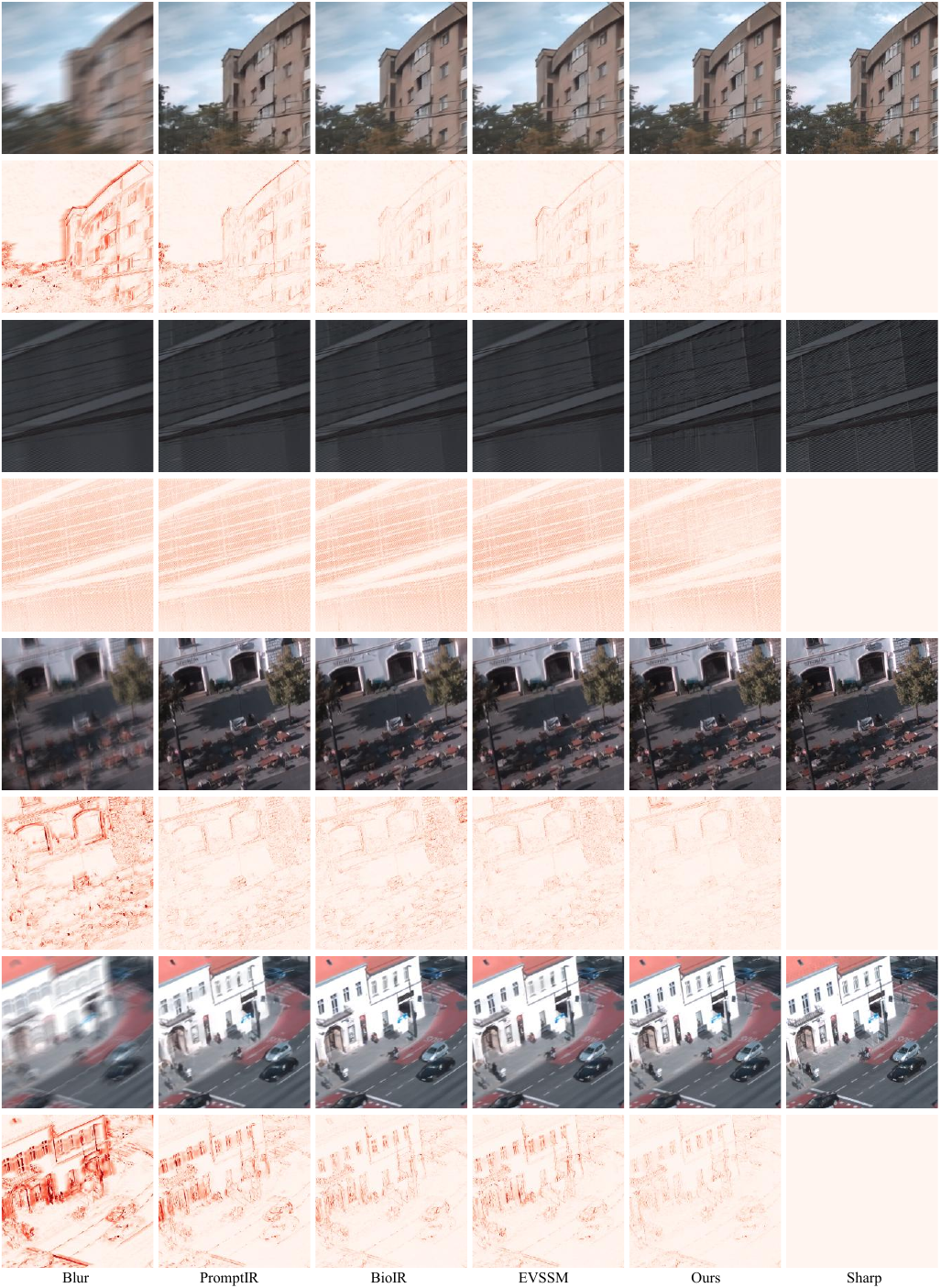}
\end{center}
\vspace{-1.3em}
\caption{Visualization comparisons on MIORe and VAR-MIORe dataset.}
\label{fig:MIORe}
\vspace{-1.em}
\end{figure*}

\begin{figure*}[t]
\begin{center}
    \includegraphics[width=1.\linewidth]{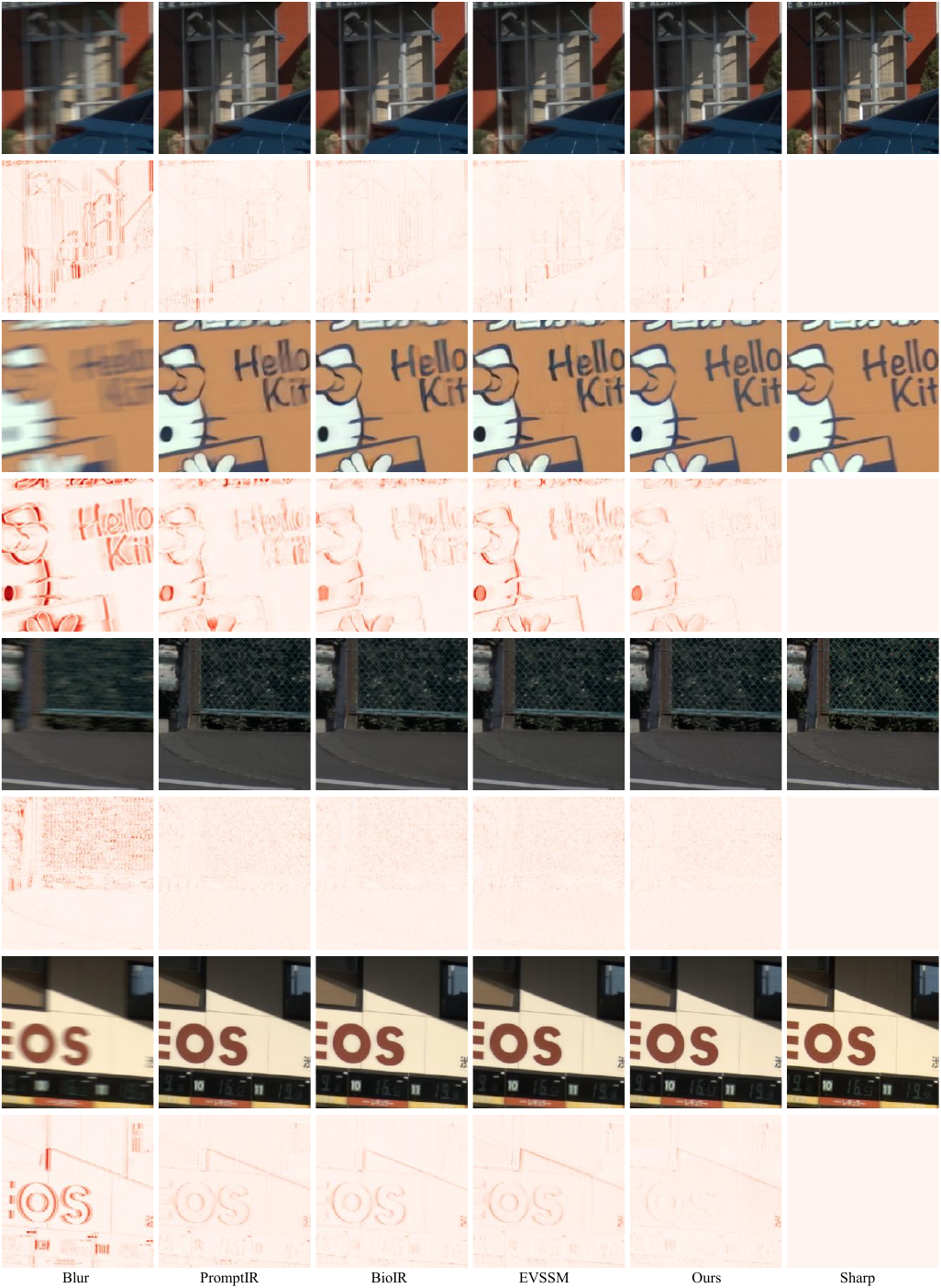}
\end{center}
\vspace{-1.3em}
\caption{Visualization comparisons on BSD dataset.}
\label{fig:BSD}
\vspace{-1.em}
\end{figure*}

\begin{figure*}[t]
\begin{center}
    \includegraphics[width=1.\linewidth]{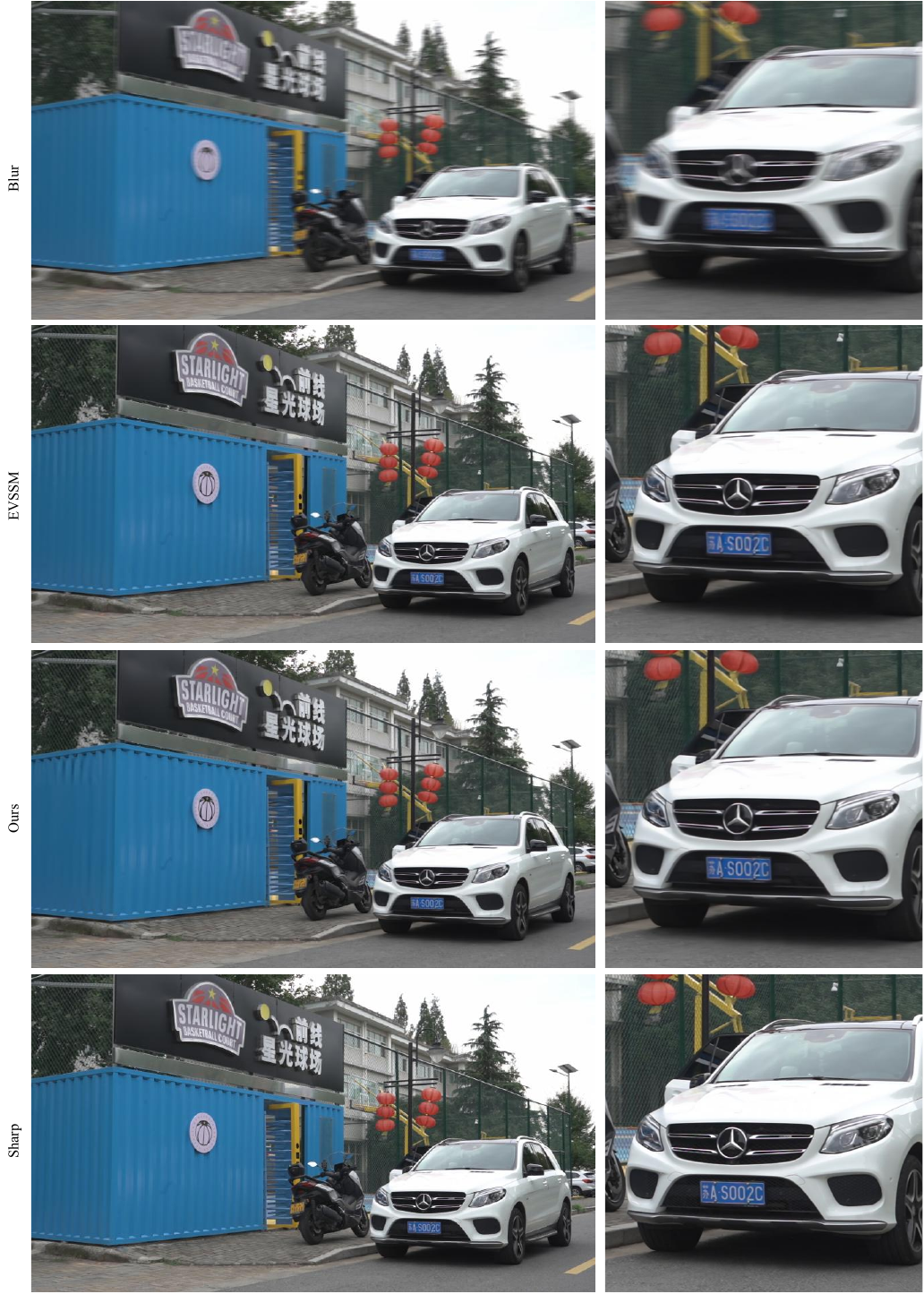}
\end{center}
\vspace{-1.3em}
\caption{Visualization comparisons on FoundIR dataset.}
\label{fig:FoundIR}
\vspace{-1.em}
\end{figure*}

\begin{figure*}[t]
\begin{center}
    \includegraphics[width=1.\linewidth]{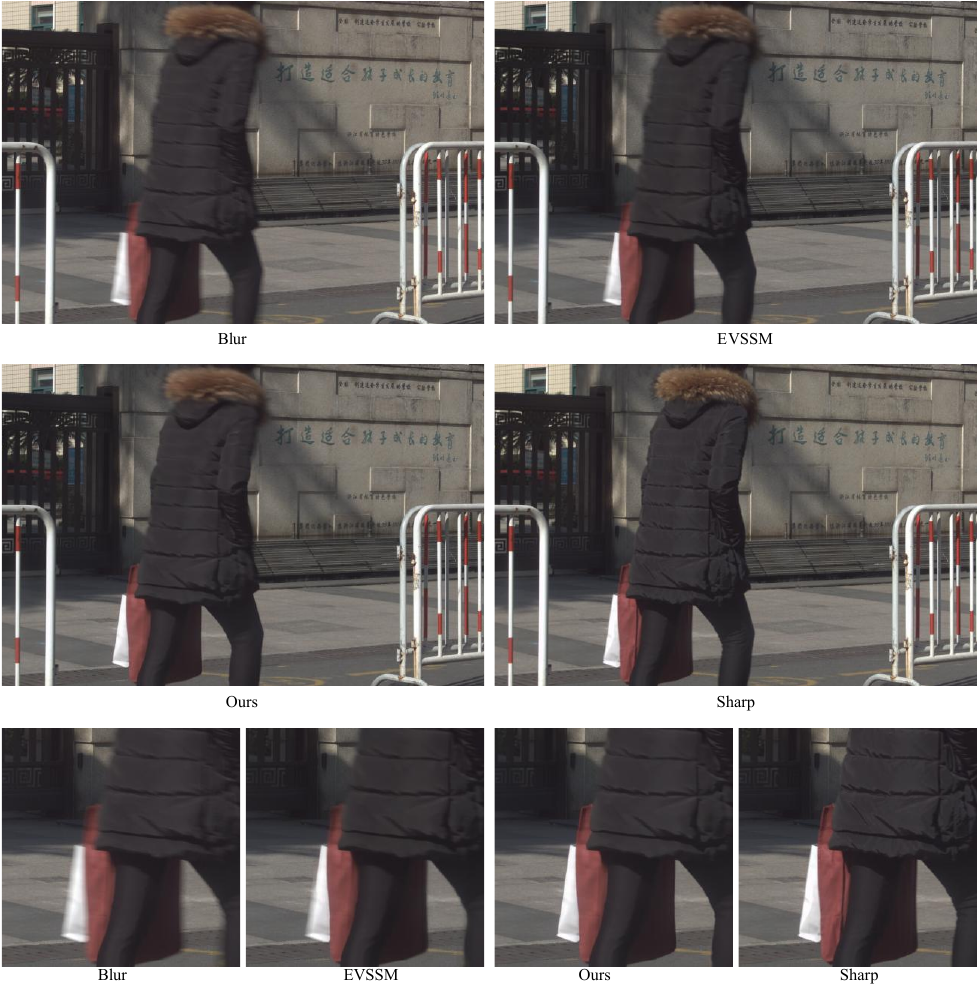}
\end{center}
\vspace{-1.3em}
\caption{Visualization comparisons on ReLoBlur dataset.}
\label{fig:ReLoBlur2}
\vspace{-1.em}
\end{figure*}

Furthermore, Fig.~\ref{fig:MIORe} $\sim$ Fig.~~\ref{fig:ReLoBlur2} illustrate the model's generalization ability on OOD test sets, such as MIORe, VAR-MIORe, BSD, FoundIR, and ReLoBlur. Specifically, in the localized close-ups from BSD (Fig.~\ref{fig:BSD}) and FoundIR (Fig.~\ref{fig:FoundIR}), DeepRFTv3-L recovers clear texts and geometric structures, yielding better visual results than the EVSSM. 

\subsection{Visualization of Removed Image Pairs}
\begin{figure*}[t]
\begin{center}
    \includegraphics[width=1.\linewidth]{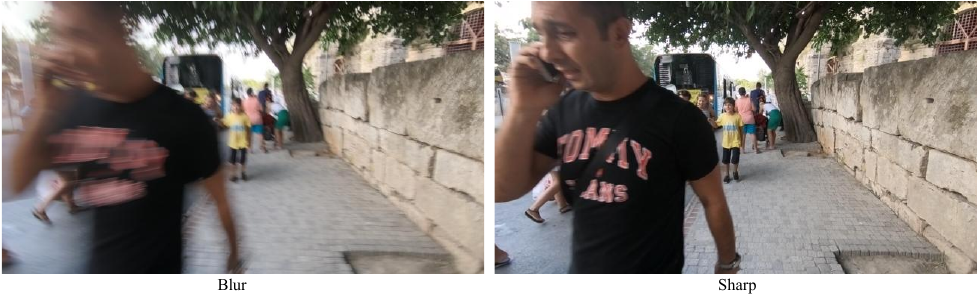}
\end{center}
\vspace{-1.3em}
\caption{Deleted image from REDS dataset.}
\label{fig:err_REDS}
\vspace{-1.em}
\end{figure*}

\begin{figure*}[t]
\begin{center}
    \includegraphics[width=1.\linewidth]{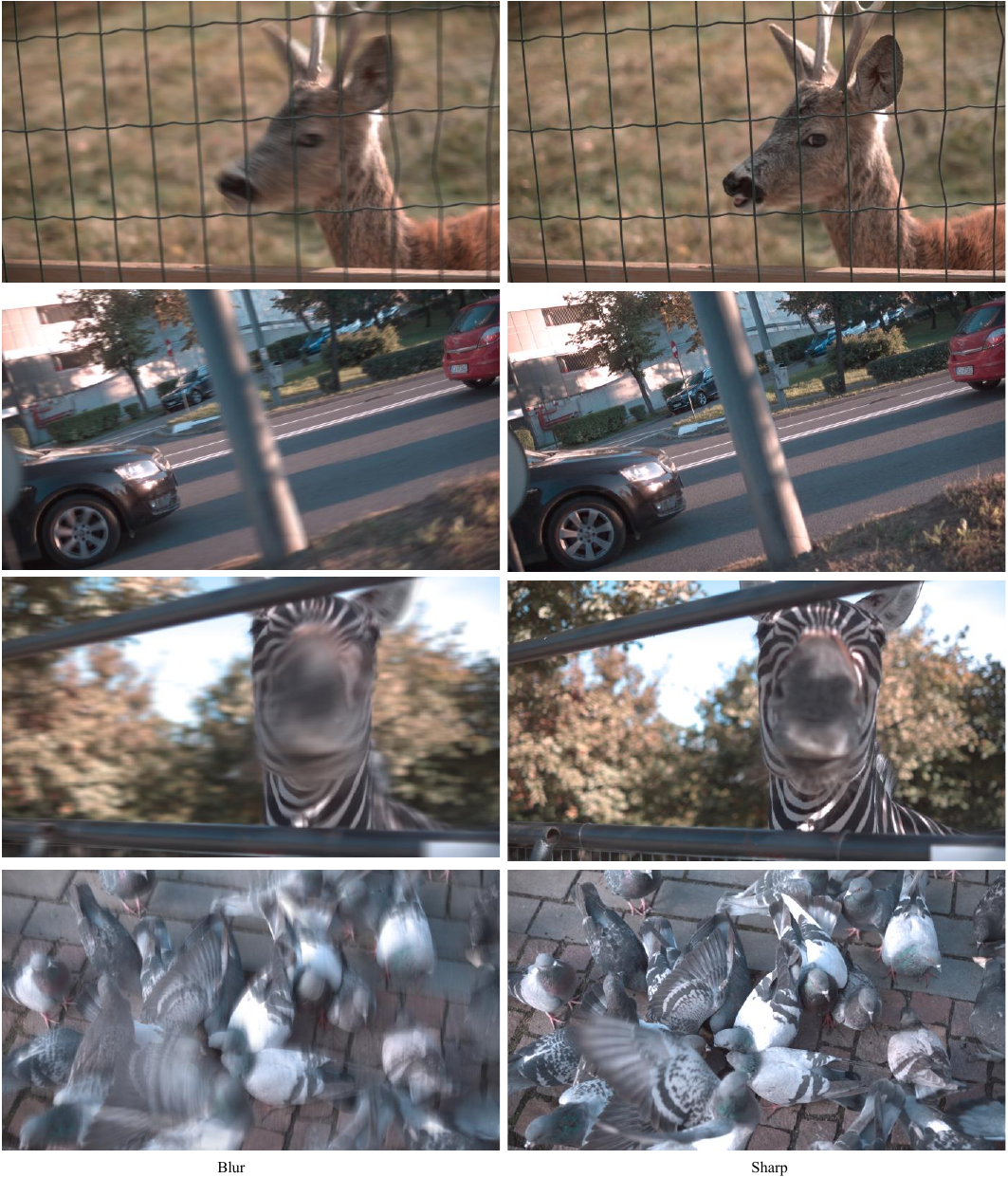}
\end{center}
\vspace{-1.3em}
\caption{Deleted image from VAR-MIORe dataset.}
\label{fig:err_VAR-MIORe}
\vspace{-1.em}
\end{figure*}

To provide a more intuitive understanding of our data pruning process, we visualize several examples of the excluded image pairs in Fig.~\ref{fig:err_REDS} $\sim$ Fig.~\ref{fig:err_VAR-MIORe}. As clearly demonstrated in these visualizations, the removed pairs suffer from severe acquisition flaws, such as residual motion blur in the ground truths and mixed degradations. This visual evidence strictly corroborates the data quality concerns discussed in the main manuscript. 

It is worth reiterating that since all models are evaluated on the exact same cleaned subsets, this unified quality control avoids evaluation bias and ensures a fair, rigorous benchmark for assessing true generalization capabilities.

\begin{figure*}[t]
\begin{center}
    \includegraphics[width=0.9\linewidth]{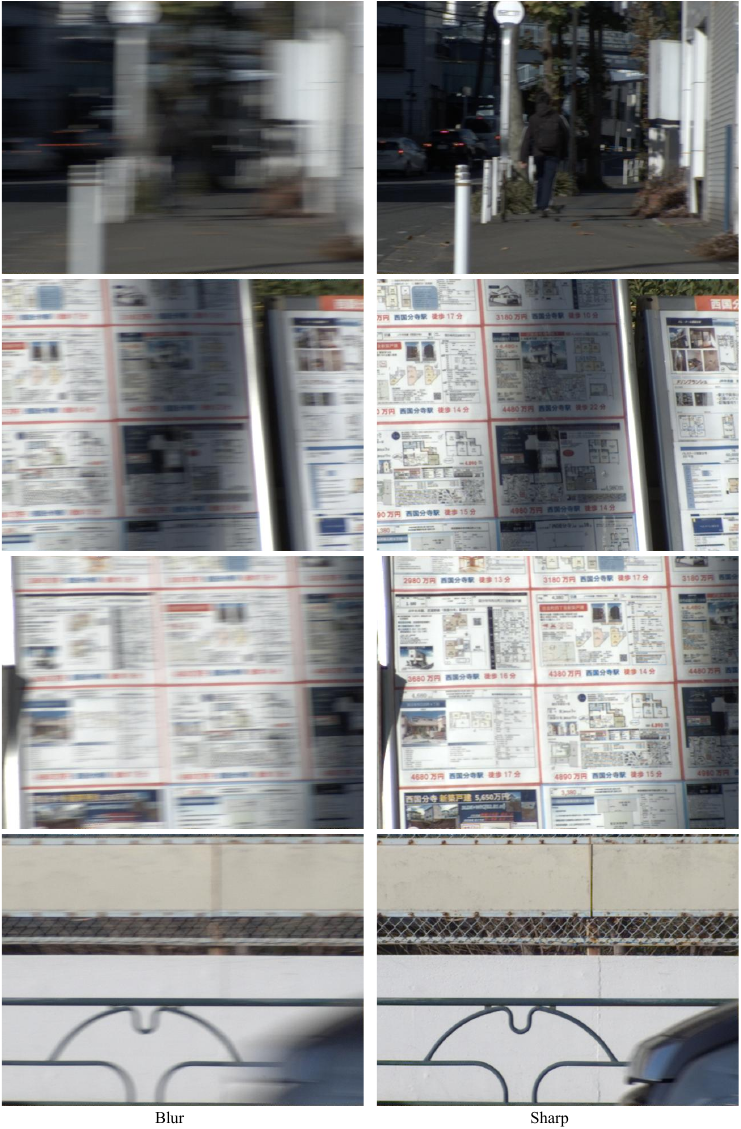}
\end{center}
\vspace{-1.3em}
\caption{Deleted image from BSD2 dataset.}
\label{fig:err_BSD2}
\vspace{-1.em}
\end{figure*}
\begin{figure*}[t]
\begin{center}
    \includegraphics[width=0.9\linewidth]{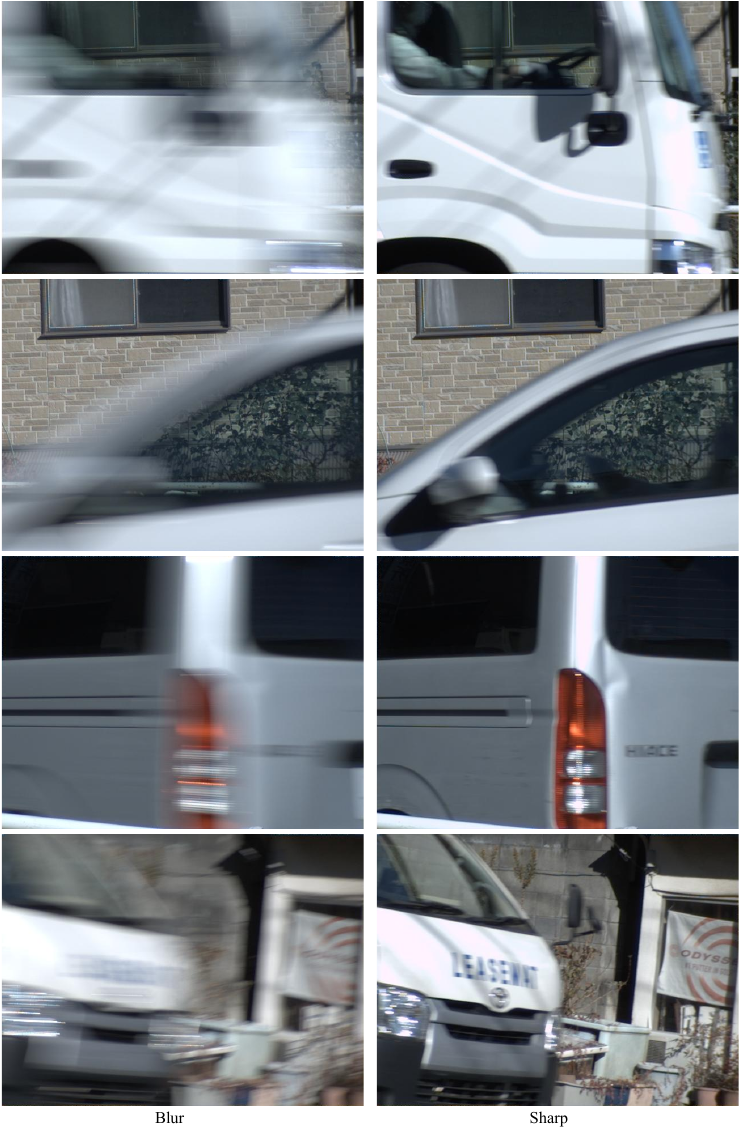}
\end{center}
\vspace{-1.3em}
\caption{Deleted image from BSD3 dataset.}
\label{fig:err_BSD3}
\vspace{-1.em}
\end{figure*}
\begin{figure*}[t]
\begin{center}
    \includegraphics[width=1.\linewidth]{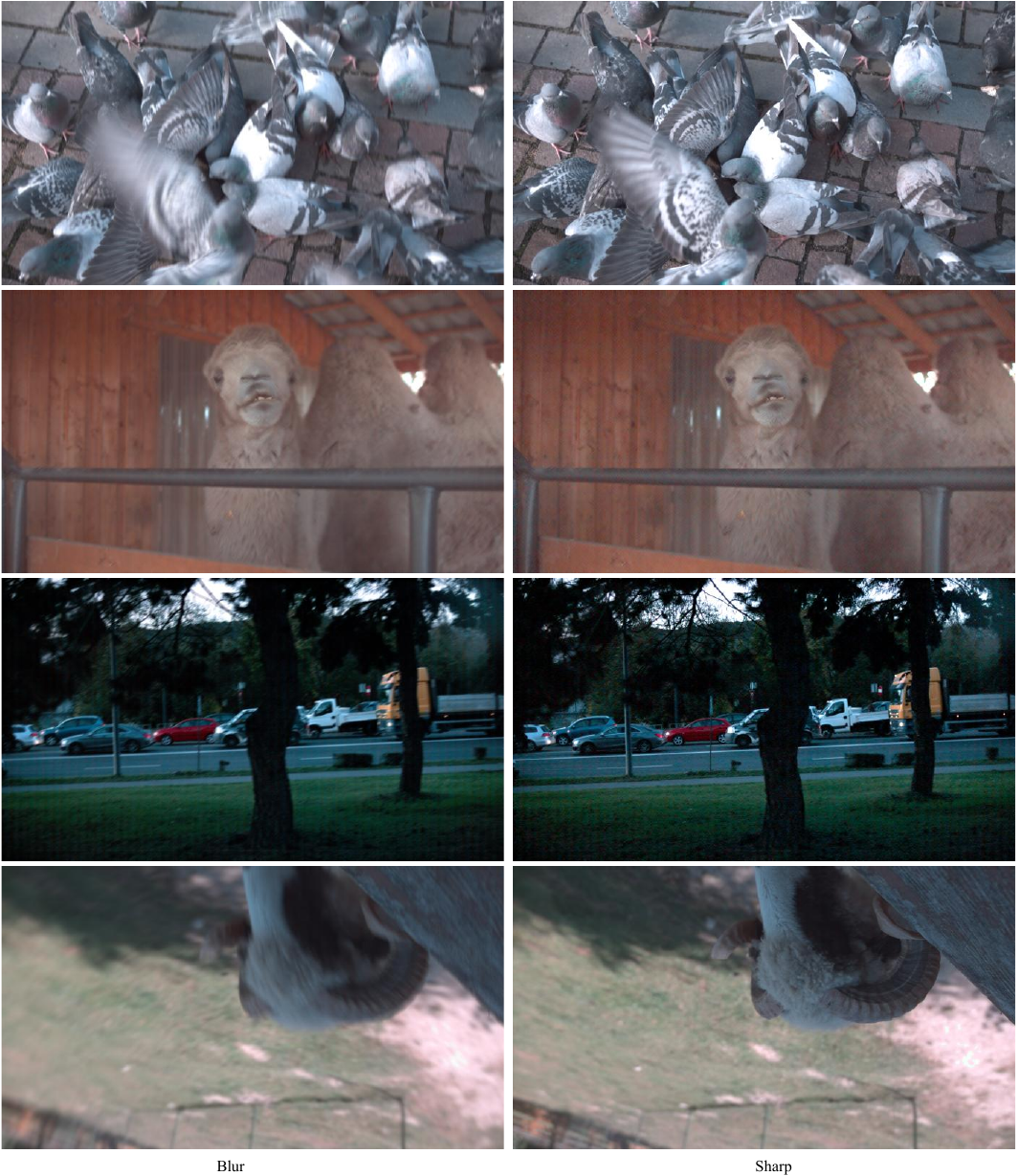}
\end{center}
\vspace{-1.3em}
\caption{Deleted image from MIORe dataset.}
\label{fig:err_MIORe}
\vspace{-1.em}
\end{figure*}

\end{document}